\documentclass[journal]{IEEEtran}
\usepackage{graphicx}
\usepackage{subfigure} 
\usepackage{threeparttable}
\usepackage{makecell}
\usepackage{array}
\usepackage{booktabs} 
\usepackage{multirow}
\usepackage{color}
\usepackage{xcolor}
\usepackage{soul}
\usepackage{cite}

\ifCLASSINFOpdf
\else
\fi
\usepackage{multirow}
\usepackage{tabularx}

\begin{document}
%
\title{Visual Semantic Segmentation Based on Few/Zero-Shot Learning: An Overview}
%
%
%

\author{Wenqi~Ren, 
	Yang Tang$^*$,~\IEEEmembership{Senior Member,~IEEE}, 
	Qiyu~Sun, 
	Chaoqiang~Zhao, 
	Qing-Long Han$^*$,~\IEEEmembership{Fellow,~IEEE}

%
\thanks{
	This work was supported by National Key Research and Development Program of China (2021YFB1714300), the National Natural Science Foundation of China (62233005), and in part by the CNPC Innovation Fund under Grant 2021D002-0902, Fundamental Research Funds for the Central Universities and Shanghai AI Lab. 
	Qiyu~Sun is sponsored by Shanghai Gaofeng and Gaoyuan Project for University Academic Program Development.  ($^*$Corresponding author: Yang Tang and Qing-Long Han.)

Wenqi Ren, Yang Tang, Qiyu Sun and Chaoqiang Zhao are with the Key Laboratory of Smart Manufacturing in Energy Chemical Process, Ministry of Education, East China University of Science and Technology, Shanghai 200237, China.  Qiyu Sun is with the Shanghai Institute of Intelligent Science and Technology, Tongji University, Shanghai, China (e-mail:  wenqiren9801@163.com; yangtang@ecust.edu.cn; qysun291@163.com; zhaocqilc@gmail.com).

Qing-Long Han is with the School of Science, Computing and Engineering Technologies, Swinburne University of Technology, Melbourne, VIC 3122, Australia (e-mail: qhan@swin.edu.au).
}}


%
%

\markboth{IEEE/CAA JOURNAL OF AUTOMATICA SINICA,~Vol.~X, No.~X, X~X}%
{Shell \MakeLowercase{\textit{et al.}}: Bare Demo of IEEEtran.cls
for Journals}
%



\maketitle

\begin{abstract}
Visual semantic segmentation aims at separating a visual sample into diverse blocks with specific semantic attributes and identifying the category for each block, and it plays a crucial role in environmental  perception. Conventional learning-based visual semantic segmentation approaches count heavily on large-scale training data with dense annotations and consistently fail to estimate accurate semantic labels for unseen categories. This obstruction spurs a craze for studying visual semantic segmentation with the assistance of few/zero-shot learning. 
The emergence and rapid progress of  few/zero-shot visual semantic segmentation make it possible to learn unseen-category from a few labeled or zero-labeled  samples, which advances the extension to practical applications. Therefore, this paper focuses on the recently published few/zero-shot visual semantic segmentation methods varying from 2D to 3D space and explores the commonalities and discrepancies of technical settlements under different segmentation circumstances. 
Specifically, the preliminaries on few/zero-shot visual semantic segmentation, including the problem definitions, typical datasets, and technical remedies, are briefly reviewed and discussed. 
Moreover, three typical instantiations are involved to uncover the interactions of few/zero-shot learning with visual semantic segmentation, including image semantic segmentation, video object segmentation, and 3D segmentation. Finally, the future challenges of few/zero-shot visual semantic segmentation are discussed. 
\end{abstract}


%
\IEEEpeerreviewmaketitle

\section{introduction}
\IEEEPARstart{V}{isual} semantic segmentation targets at fine-grained classification for collected samples, such as images, videos, and 3D meshes. It has gradually drawn research interests in virtue of the extensive applications to self-driving \cite{chen2018parallel,sun2020proximity}, medical diagnosis \cite{he2022differentiable,valanarasu2021medical}, remote sensing \cite{cheng2022multi}, 
and so on. As three typical representatives in visual semantic segmentation, image semantic segmentation (ISS) \cite{multitask,cheng2021per}, video object segmentation (VOS) \cite{patil2021unified,garg2021mask} and 3D segmentation (3DS) \cite{hu2021learning,te2018rgcnn} have been well-developed. Among them, ISS and VOS attempt to automatically assign a label for each pixel located in  2D images, while 3DS  endeavors to allocate annotations to target shapes or objects for given 3D samples, such as point clouds \cite{hu2021learning,te2018rgcnn} and 3D meshes \cite{hou20213d}. 

\begin{figure}[!t]
	\centering
	\includegraphics[width=\linewidth]{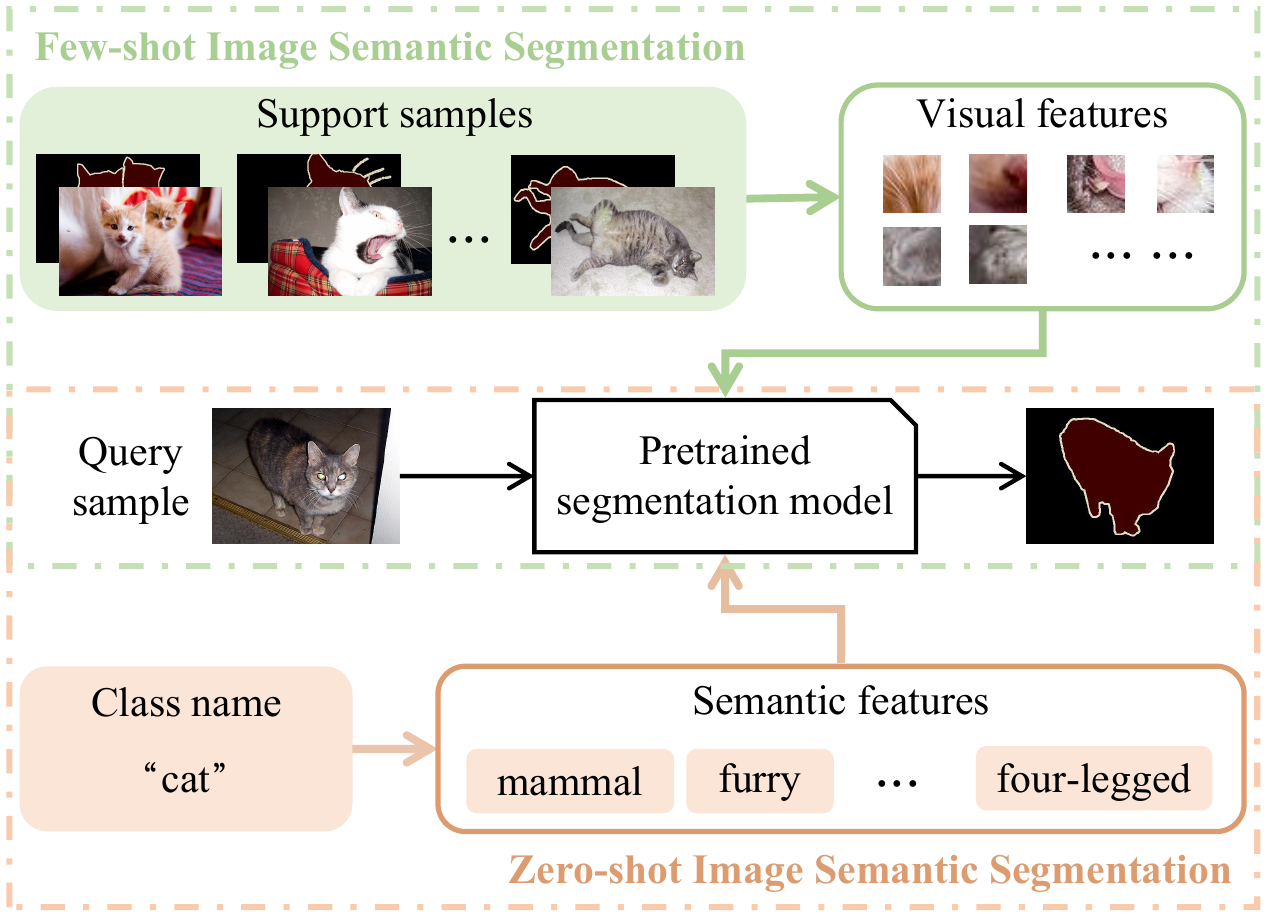}
	\caption{An example of  few/zero-shot visual semantic segmentation, where \emph{the segmentation model is pre-trained on the categories except ``cat''}. The visual samples are selected from the PASCAL VOC \cite{everingham2010pascal}. }
	\label{supervised}
\end{figure}

\begin{table*}[!t]
	\centering
	\begin{threeparttable}[b]
		\caption{A summary of technical solutions for few/zero-shot visual semantic segmentation tasks}
		\renewcommand\arraystretch{1.15}
		\label{summ}
		\begin{tabular}{c|p{13mm}<{\centering}|c|p{45mm}<{\centering}|p{17mm}<{\centering}|p{27mm}<{\centering}|c}
			\hline
			\multirow{1.5}{*}{\textbf{Space}} & \multirow{1.5}{*}{\textbf{Samples}}  & \multirow{1.5}{*}{\textbf{Scenarios}}  & \multirow{1.5}{*}{\textbf{Profiles}}   & \multirow{1.5}{*}{\textbf{Solutions}} & \multirow{1.5}{*}{\textbf{Typical datasets}}& \multirow{1.5}{*}{\textbf{Literatures}} \\ \hline
			\multirow{11}{*}{2D}&\multirow{6}{*}{Images}&\multirow{4}{*}{Few-shot ISS}&\multirow{4}{45mm}{Predict pixel-wise labels for unseen  categories with a few annotated images} & Metric& \multirow{4}{27mm}{PASCAL-5$^i$ \cite{shaban2017one}, COCO-20$^i$ \cite{nguyen2019feature}}&\cite{dong2018few,wang2019panet,zhang2019canet,lang2022learning,rakelly2018conditional,tian2020prior,chen2021semantically,hu2019attention,yang2020new,zhang2020sg,liu2022axial,yang2021part,li2021adaptive,liu2020part,zhang2021rich,wang2021variational,zhang2019pyramid,wang2020few,liu2021harmonic,zhang2021few,gao2022mutually,shi2022dense}\\
			&&                                                     &                         & Parameter&&\cite{shaban2017one,zhuge2021deep,liu2020dynamic}\\
			&&                                                     &                         & Fine-tune&&\cite{yang2020brinet,tian2020differentiable,boudiaf2021few}\\
			&&                                                     &                         & Memory&&\cite{wu2021learning,xie2021few}\\ \cline{3-7}
			&&\multirow{2}{*}{Zero-shot ISS}&\multirow{2}{45mm}{Predict pixel-wise labels for unseen categories with zero annotated images} & Metric& \multirow{2}{27mm}{PASCAL VOC \cite{everingham2010pascal}, PASCAL Context \cite{mottaghi2014role}}&\cite{xian2019semantic,tian2020cap2seg,lu2021feature,hu2020uncertainty,kato2019zero,baek2021exploiting}\\
			&&                                                     &                         & Generative&&\cite{bucher2019zero,gu2020context,gu2020pixel,li2020consistent}\\ \cline{2-7}
			&\multirow{5}{*}{Videos}&\multirow{3}{*}{Few-shot VOS}&\multirow{3}{45mm}{Segment the object specified in the first frame in the remaining frames} & Metric& \multirow{3}{27mm}{Youtube-VOS \cite{xu2018youtube}, DAVIS 2016 \cite{Perazzi2016}}&\cite{wang2019fast,yin2021directional,motionguide,voigtlaender2019feelvos,yang2020collaborative,zhu2021separable,zhang2020transductive,liu2020guided,hong2021adaptive,park2021learning}\\
			&&                                                     &                         & Fine-tune&&\cite{khoreva2019lucid,robinson2020learning,xiao2019online,meinhardt2020make,xu2021meta}\\
			&&                                                     &                         & Memory&&\cite{oh2019video,oh2020space,seong2020kernelized,seong2022video,xie2021efficient,seong2021hierarchical,li2020fast,liang2020video,cheng2021rethinking,lu2020video,zhang2020dual,lyu2019lip}\\ \cline{3-7}
			&&\multirow{3}{*}{Zero-shot VOS}&\multirow{3}{45mm}{Segment the primary object in the video sequence without annotations} & Metric&\multirow{3}{27mm}{Youtube-VOS \cite{xu2018youtube}, DAVIS 2016 \cite{Perazzi2016}}&\cite{yang2019anchor,liu2021f2net,gu2020pyramid,lu2020zero,zhuo2019unsupervised,zhou2020motion,zhou2021flow,yang2021learning,ji2021full,zhao2021multi,seo2020urvos,bellver2020refvos,yang2021hierarchical,botach2022end}\\
			&&&&Fine-tune&&\cite{li2022you}\\
			&&                                                     &                         & Memory&&\cite{ventura2019rvos,tokmakov2019learning,mahadevan2020making,wang2019zero}\\ \cline{1-7}
			\multirow{5}{*}{3D}&\multirow{5}{15mm}{Point clouds /3D meshes}&\multirow{3}{*}{Few-shot 3DS}&\multirow{3}{45mm}{Assign labels for unseen categories with a few annotated 3D samples} & Metric& \multirow{3}{27mm}{ShapeNet  Part \cite{Fan_2017_CVPR}, ScanNet \cite{dai2017scannet}}&\cite{chen2020compositional,zhao2021few,yuan2020ross,wang2020few1}\\
			& &                                                    &                         & Parameter&&\cite{hao20213d}\\
			& &                                                    &                         & Fine-tune&&\cite{huang20213d,sharma2019learning,sharma2020self,chen2019bae}\\ \cline{3-7}
			&&\multirow{2}{*}{Zero-shot 3DS}&\multirow{2}{45mm}{Assign labels for unseen categories with zero annotated 3D samples} & \multirow{2}{*}{Generative}&\multirow{2}{27mm}{ScanNet \cite{dai2017scannet}, S3DIS \cite{armeni20163d}}&\multirow{2}{*}{\cite{michele2021generative,liu2021segmenting}}\\ 
			&&&& &\\ \hline
		\end{tabular}
		\begin{tablenotes}
			\item[1]  Each technical solution is represented by its first word.
		\end{tablenotes}
	\end{threeparttable}
\end{table*}

Thanks to the advance in deep learning, the conventional supervised learning-based visual semantic segmentation approaches have made considerable breakthroughs in both real-time performance and prediction accuracy \cite{hong2021deep,yu2021bisenet,guo2020deep,liu2021fiss}. They learn to fit the distribution of specific classes  from a great deal of training samples and strive to segment them on test samples. 
Whereas, these visual semantic segmentation algorithms still confront various challenges. Firstly, a great deal of training data with fine-grained annotations is desperately required for these segmentation models to learn a category of interest, where the annotations are expensive to obtain. Particularly, in 3D point cloud segmentation, the order of millions is commonly reached by the number of captured point clouds \cite{cheng2021sspc}, leading to laborious labeling processes. Moreover, the structure of point clouds is irregular, which further attaches enormous challenges to manual annotations. 
Secondly, these algorithms rely heavily on an implicit assumption that the test categories  must be the same as the training ones \cite{goodfellow2016deep,lecun2015deep,sunsurvey}. Consequently, when dealing with unseen categories after training,  these segmentation  models always suffer from severe performance degradation, which hinders their flexible application in dynamic scenarios.  
Although the methods based on weakly-supervised learning \cite{zhang2021weakly,kim2021discriminative,lee2021reducing} alleviate the data hunger significantly, they still require that  the categories encountered at test-time must have been seen  in training, leading to the problem of cross-class adaptation failing to be  addressed. 
These limitations encourage the employment of few-shot learning and zero-shot learning \cite{wang2020generalizing,wang2019survey,hu2022can} on visual semantic segmentation, which promotes the skills of segmentation models in dealing with unseen categories, even though they have only been exposed to a few labeled, or even unlabeled,  samples, as shown in Fig. \ref{supervised}.

\begin{figure}[!t]
	\centering
	\includegraphics[width=0.9\linewidth]{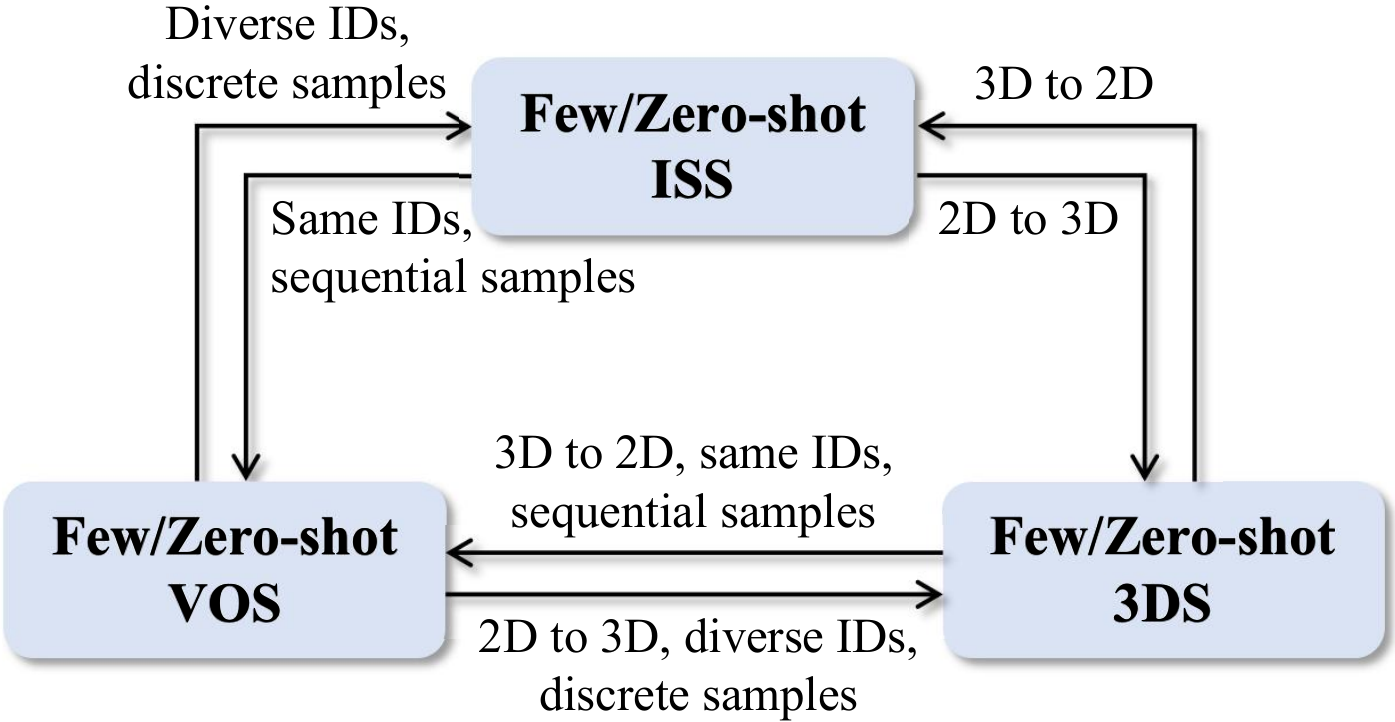}
	\caption{The relationship among ISS, VOS, and 3DS under both few-shot and zero-shot cases.}
	\label{relationship}
\end{figure}

In recent years, a significant number of meaningful and valuable few/zero-shot visual semantic segmentation studies \cite{sun2021metaseg,yao2020video,wang2021survey}  have been spawned, which eliminates the barriers on the cross-class adaptation  with a limited number of annotated samples. To summarize these approaches, some related and well-written surveys were published \cite{sun2021metaseg,yao2020video,wang2021survey}.  Compared with these previous efforts \cite{sun2021metaseg,yao2020video,wang2021survey}, this paper creatively summarizes  relevant studies from the perspective of problem settings and technical solutions of few/zero-shot learning and strengthens the relationship between different segmentation methods, including ISS, VOS, and 3DS, in both few-shot and zero-shot scenarios. In addition, this paper provides some discussions on open challenges that few/zero-shot learning brought to visual semantic segmentation, such as cross-domain few/zero-shot segmentation and generalized few/zero-shot segmentation. In summay, the main contributions of this paper are as follows:

\begin{figure*}[!t]
	\centering
		\includegraphics[width=\linewidth,height=7cm]{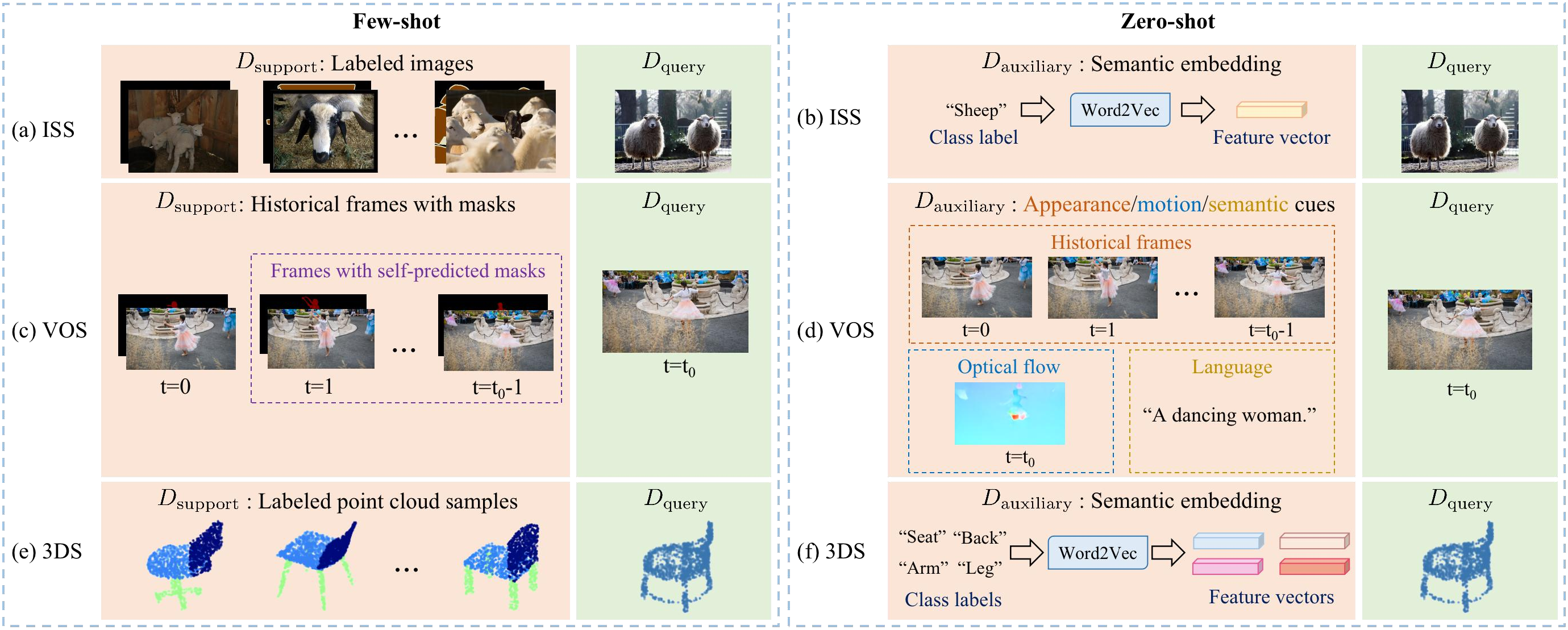}
	\caption{The conventional settings of different few/zero-shot visual segmentation tasks. 
			The ISS samples are selected from the PASCAL VOC \cite{everingham2010pascal}, the VOS samples are picked from DAVIS 2016 \cite{Perazzi2016}, and the 3DS exemplars are sampled from  ShapeNet Part \cite{Fan_2017_CVPR}. The Word2Vec \cite{mikolov2013efficient,wang2016relational} is the operation to map a class label as a feature vector, which is called word embedding \cite{wang2019survey} and represents the semantic attributes of the corresponding category. }
\label{support}
\end{figure*}

\begin{itemize}
\item The comparison among the problem settings of  different few/zero-shot visual segmentation tasks and  a summary of technical solutions are provided. 

\item The advancements of few/zero-shot visual semantic segmentation are reviewed and  the dissimilarities of technical solutions in different segmentation tasks are specified.

\item The open challenges that involve data, algorithms, and applications  for few/zero-shot  visual semantic segmentation are discussed to  provide the enlightenment to follow-up researchers.
\end{itemize}

The rest of this paper is organized as follows. Section \ref{section2} provides the comparison on the problem settings and typical datasets of  different few/zero-shot visual segmentation tasks and gives a summary of technical solutions.  Section  \ref{section5}, Section \ref{section6} and Section \ref{section7} concentrate  on the advancements of ISS, VOS and  3DS in both few-shot and zero-shot circumstances, respectively, and illustrate the technical settlements in diverse segmentation scenarios. 
Section \ref{section8} analyzes the open challenges and applications for few/zero-shot visual semantic segmentation.

\section{preliminaries}\label{section2}

Few-shot learning expects to learn unseen categories from a few  labeled training samples \cite{wang2020generalizing,hu2022can,parnami2022learning}. When the annotated samples are unavailable, the few-shot learning problem turns into a zero-shot learning problem  \cite{wang2020generalizing,parnami2022learning}.  Therefore, zero-shot learning is the boundary case of few-shot learning \cite{wang2020generalizing,antonelli2021few,song2022comprehensive}.  
This section attempts to  provide  detailed definitions, typical datasets, and technical remedies  for  few/zero-shot visual semantic segmentation.

\subsection{Problem Definition}

As shown in Table \ref{summ}, few/zero-shot ISS methods  \cite{dong2018few,wang2019panet,li2020consistent} strive to learn a powerful segmentation model that can estimate pixel-wise labels for unseen objects, even though they have only been exposed to a small number of labeled, or even unlabeled, images. When learning a specific unseen category with a few or zero annotated images, there is no constraint on the identities of the objects in training and reasoning samples. Few/zero-shot VOS \cite{wang2019fast,mahadevan2020making,wang2019zero}, which aims at segmenting foreground objects under the guidance of a limited number of labeled frames, can be regarded as a particular case of  ISS \cite{dong2018few,wang2019panet,li2020consistent}, as the images segmented here  are continuous in time steps, and  the target objects generally have the same  identities but different appearances  in sequential frames. Few/zero-shot 3DS \cite{chen2020compositional,zhao2021few,liu2021segmenting} conducts segmentation in  3D space rather than 2D space, where the inputs are disordered 3D samples, such as point clouds and meshes, rather than regular 2D images. The settings and  relationship of them are illustrated in Fig. \ref{relationship} and Fig. \ref{support}, respectively.

\begin{table}[!t]
	\centering
	\caption{The nomenclature of the symbolism involved in  this paper}
	\label{meaning}
			\renewcommand\arraystretch{1.15}
	\begin{tabular}{p{20mm}<{\centering}p{50mm}<{\centering}}
		\hline
		\textbf{Symbolism} & \textbf{Meaning} \\ \hline
		$T$ & Few-shot/zero-shot task\\
		$x$ & Visual samples\\
		$m$ & Annotations \\
		$w$ & Category description \\
		$e$ & Word embeddings \\
		$n$ & Noise \\
		$\omega$ & Weights of the prediction layer \\
		$\hat{m}$ &Prediction of $x$\\ 
		$\hat{z}$ &Prediction of $e$\\ 
		$\mathcal{X}_T$ & Input space of unseen classes \\
		$\mathcal{M}_T$ & Label space of unseen classes \\
		$\mathcal{X}_S$ & Input space of base classes \\
		$\mathcal{M}_S$ & Label space of base classes \\
		$D_{\rm{support}}$ &  Support set  \\
		$D_{\rm{query}}$ &  Query set  \\
		$D_{\rm{source}}$ & Dataset of base classes \\
		$D_{\rm{auxiliary}}$ & Auxiliary dataset \\
		$I_{\rm{support}}$ & Number of samples in $D_{\rm{support}}$\\
		$I_{\rm{query}}$ & Number of samples in $D_{\rm{query}}$\\
		$I_{\rm{source}}$ & Number of samples in $D_{\rm{source}}$\\
		\hline
	\end{tabular}
\end{table}

In this section, we provide a problem definition of visual semantic segmentation in both few-shot and zero-shot scenarios. In each circumstance,  we take ISS as an instance and deliver a detailed definition. In addition, we discuss the differences among the settings of few-shot ISS, few-shot VOS, and few-shot 3DS, and extend these problem settings into zero-shot scenarios.  
The nomenclature of the symbolism involved in  this paper is listed in Table \ref{meaning}.

\subsubsection{Few-shot Visual Segmentation} 
The problem definition of few-shot ISS is firstly discussed, which will be further extended to few-shot VOS and few-shot 3DS. 
Few-shot ISS aims to learn a segmentation model from a few labeled images for target categories, as illustrated in Table \ref{summ}. A few-shot ISS task $T$ attempts to learn a powerful segmentation model that is capable of predicting the pixel-wise  semantic category from  $D_{\rm{support}}$ for  images in  $D_{\rm{query}}$. The $D_{\rm{support}}={\{(x_i,m_i)\}}_{i=1}^{I_{\rm{support}}}$ called the support set denotes a dataset containing $I_{\rm{support}}$ images $x$ with their corresponding labels $m$, and $D_{\rm{query}}={\{(x_k)\}}_{k=1}^{I_{\rm{query}}}$ termed the query set represents a dataset composed of $I_{\rm{query}}$ unlabeled images.  The images from both $D_{\rm{support}}$ and $D_{\rm{query}}$ have the same marginal distribution $P_{\mathcal{X}_T}$ of the input space $\mathcal{X}_T$ and the same label space $\mathcal{M}_T$. 
Typically, the $I_{\rm{support}}$ is set to $CK$, where the $I_{\rm{support}}$ samples consists of  $C$ classes, and each  class provides $K$ samples (generally $K$ is set to 1 or 5 \cite{wu2021learning}). In these circumstances, the  task $T$ is allowed to be termed as a $C$-way $K$-shot task. Fig. \ref{support}(a) indicates an example of a few-shot ISS task, where the category of both $D_{\rm{support}}$ and $D_{\rm{query}}$ is  ``sheep". 

Whereas, directly adopting the conventional supervised learning to reap a satisfactory segmentation model  from insufficient data $D_{\rm{support}}$  is an exceptionally ambitious and challenging attempt. As a consequence, most of  few-shot ISS approaches resort to the assistance of prior knowledge, which is distilled from an accessible dataset $D_{\rm{source}}={\{(x_j,m_j)\}}_{j=1}^{I_{\rm{source}}}$ (if any) and is conducive to dealing with the target  $T$. The source dataset $D_{\rm{source}}$ contains $I_{\rm{source}}$ labeled samples for base classes $(I_{\rm{source}}\gg I_{\rm{support}})$ and shares the same input space $\mathcal{X}_S=\mathcal{X}_T$ but distinct label space $\mathcal{M}_S$ with $D_{\rm{support}}$ and $D_{\rm{query}}$. By combining the available supervision information of  $T$ with the learned prior knowledge,  obtaining a high-quality  model becomes more feasible and reasonable \cite{wang2020generalizing}. 

The settings of few-shot VOS and few-shot 3DS follow that of few-shot ISS, but also preserve their own distinct identities. The samples to be processed in few-shot VOS are sequential  frames. Given only the first frame of a video with annotations, few-shot VOS strives to predict pixel-wise labels for the specific objects in the remaining frames. Few-shot VOS is a sequential learning problem that segments moving objects with shared identities frame by frame. Since the appearance of the objects changes dramatically between different frames, the $D_{\rm{support}}$ in few-shot VOS are allowed to be comprised of the first frame with given labels and/or the historical frames with self-predicted masks to capture reliable object characteristics. For few-shot 3DS, it can be observed in Fig. \ref{support}(c) that the type of inputs is generally constructed by 3D exemplars.  Take few-shot point cloud part segmentation for example. The $D_{\rm{support}}$  is made of point cloud samples whose subparts are annotated by  unseen categories, and the $D_{\rm{query}}$  is composed of the ones with subparts of the same categories as the $D_{\rm{support}}$. 

\subsubsection{Zero-shot Visual Segmentation} 
When ISS encounters zero-shot scenarios, the task $T$ will become more complex  to address. In this circumstance, the  $D_{\rm{support}}$ fails to access any images with supervision signals, which hinders the learning on the target  $T$. 
To address this issue, zero-shot ISS ordinarily attempts to transfer some supervision information from other modalities, denoting as the auxiliary dataset $D_{\rm{auxiliary}}$, to enable the learning practicable, as depicted in Fig. \ref{supervised}. 
Specifically, 
auxiliary information \cite{wang2019survey} from semantic embeddings  is applied to tackle the zero-shot ISS task $T$, as shown in Fig. \ref{support}(d). The dataset  $D_{\rm{auxiliary}}=\{(w_j, e_j)\}_{j=1}^{N}$ contains  $N$ categories, where $w_j$ and $e_j$ represent the category description  and the word embedding  corresponding to the $j$-th specific class, respectively. Thanks to  $D_{\rm{auxiliary}}$, it is hopeful for zero-label unseen objects to be separated effectively.

As a special case of few-shot VOS,   the primary target object in the video sequence is demanded to be segmented without accessible annotations in zero-shot VOS. Since there is abundant  temporal information that is  hidden in the sequential frames or  optical flows, zero-shot VOS often exploits appearance and motion cues  to provide adequate supervision signals for task $T$. Moreover, language expressions can also supervise the model to segment objects of interest, as illustrated in Fig. \ref{support}(d).  It can be seen in Fig. \ref{support}(f) that zero-shot 3DS follows the scheme of zero-shot ISS, which also uses word embeddings as crucial sources of auxiliary supervision signals. However,  the target samples in zero-shot 3DS are 3D visual data rather than 2D images, which is distinguished from zero-shot ISS. 

\subsection{Typical Datasets}

In this section, two typical datasets of  each visual semantic segmentation scenario are selected to be  discussed.  The picked datasets are displayed in Table \ref{summ}. 

\emph{PASCAL-5$^i$} \cite{shaban2017one} is a popular benchmark designed for few-shot ISS, which is created by PASCAL VOC 2012 \cite{everingham2010pascal} with additional annotations from SDS dataset \cite{hariharan2014simultaneous}. There are 20 categories, split into 4 subsets with 5 classes per subset. 

\emph{COCO-20$^i$}  \cite{nguyen2019feature} is the largest dataset for few-shot ISS, which is built from MS COCO benchmark \cite{lin2014microsoft}. It covers 80 common categories, which are divided into 4 folds with 20 categories per fold. 

\emph{PASCAL VOC}   \cite{everingham2010pascal} can be adopted for image classification, object detection, and object semantic segmentation. For segmentation tasks, there are 1464 available samples for training and 1449 samples for validation with 20 categories in total. 

\emph{PASCAL Context} \cite{mottaghi2014role} is proposed for scene parsing, which covers both indoor and outdoor scenarios. It contains 4998 training and 5105 validation samples with 59 classes. 
 
\emph{Youtube-VOS} \cite{xu2018youtube} is presented for VOS tasks, which has 94 classes in total. It consists of 3471 videos with 65 seen categories for training, 507 videos with 65 seen and 26 unseen categories for validation, and 541 videos with 65 seen and 29 unseen categories for testing. 

\emph{DAVIS 2016} \cite{Perazzi2016} is designed for segmenting fore- and background objects in videos, where only  binary annotations are provided. There are 30 videos for training and 20 videos for validation. 

\emph{ShapeNet  Part} \cite{Fan_2017_CVPR}  is applied for 3D fine-grained point cloud segmentation. It comprises  16881 3D shape instances of 16 classes.  Each instance is further labeled by 2-5 part annotations, leading to 50 part categories. 

\emph{ScanNet} \cite{dai2017scannet} is a 3D point cloud dataset captured by RGB-D cameras, which contains 1513 diverse indoor scenarios  (1201 for training and 312 for testing) with 20 classes. 

\emph{S3DIS} \cite{armeni20163d} is a dataset for 3D semantic segmentation, which includes RGB-D data and 3D point clouds. The 3D point clouds are scanned from 271 indoor environments with 13 categories. 

\begin{figure*}[!t]
	\centering
	\includegraphics[width=\linewidth]{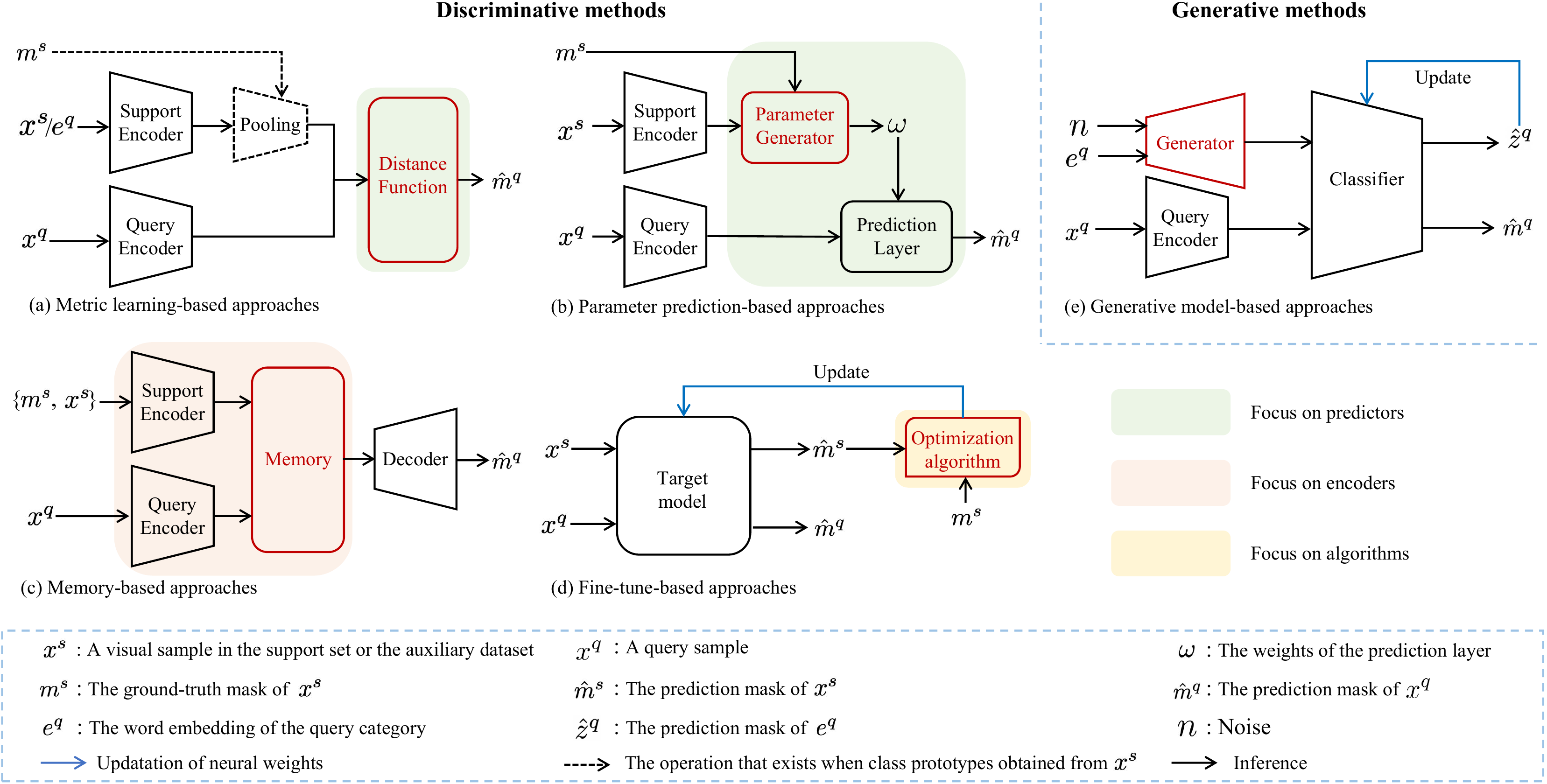}
	\caption{Conventional frameworks of few/zero-shot visual segmentation. The principal differences of these  frameworks are highlighted in red.}
\label{typical}
\end{figure*}

\subsection{Technical Solutions}
 Given a query image $x_k$, the existing few/zero-shot visual segmentation approaches strive to estimate the  posterior probability of classes the pixels in $x_k$ belong to. For convenience, we describe this as follows: 
\begin{equation} 
	\label{eq1}
		\hat{m}_k=\arg \max P(m_k|x_k). 	
\end{equation}
We further consider the Bayesian rule, then the above equation can be expressed as:
\begin{equation}
	\label{eq2}
	\hat{m}_k=\arg \max P(x_k|m_k)P(m_k), 
\end{equation}
where  $P(x_k|m_k)$ is the  the conditional distribution of $x_k$ given the mask $m_k$, and $P(m_k)$ is the  prior distribution of $m_k$.

Based on Eq. \ref{eq1} and Eq. \ref{eq2}, we can summarize the few/zero-shot visual segmentation approaches into discriminative approaches and generative approaches, where the discriminative approaches attempt to build their frameworks to maximize  $P(m|x)$ and the generative approaches strive to model $P(x|m)P(m)$. Thus, discriminative methods expect to construct  a powerful segmentation model so that it can maximize the posterior probability of classes on query samples. Discriminative methods tend to capture a more effective mask predictor, grasp an optimization algorithm with fast convergence, or enhance the representations of objects, leading to four sub-categories: metric learning-based, parameter prediction-based, fine-tune-based, and memory-based methods. The conventional frameworks of the above four technical solutions are exhibited in Fig. \ref{typical}(a)-(d). 
 As for the generative approaches, a generator  is frequently built  to  fit the distribution of $x$ or its features conditioned on classes in $m$, where $P(m)$ is generally supposed to be a uniform distribution, as shown in Fig. \ref{typical}(e). 
In the rest of this section, we will describe these five types of technical solutions in detail. 

\subsubsection{Metric Learning-based Methods}
Metric learning-based approaches attempt to learn a general distance function so that it can provide higher affinity scores to more similar features and lower scores to more distinct features   for any category.  
The distance functions  can be any algorithms or networks so long as they can calculate the distance between two representations.  
Since metric learning-based methods achieve segmentation through calculating pairwise similarity, the structures of few/zero-shot tasks, such as the number of categories, would not be strictly  required \cite{lu2020learning1}, making them more flexible.  Nevertheless, the high correlation between the source and target tasks is a potential condition for metric learning-based methods to be well-implemented \cite{wang2020generalizing}.

 \begin{table*}[!t]
	\centering
	\begin{threeparttable}[b]
		\caption{A summary on common advantages and disadvantages of the five few-shot solutions.}
		\renewcommand\arraystretch{1.25}
		\label{ad}
		\begin{tabular}{cc|m{5cm}|p{10mm}<{\centering}p{10mm}<{\centering}|p{10mm}<{\centering}p{10mm}<{\centering}|p{10mm}<{\centering}p{10mm}<{\centering}}
			\hline
			\multicolumn{2}{c|}{\multirow{3}{*}{\textbf{Type of Methods}}} & \multicolumn{1}{c|}{\multirow{3}{*}{\makecell[c]{\textbf{Characteristics}}}}  &  \multicolumn{6}{c}{ \makecell[c]{\textbf{Method performance}}}  \\ \cline{4-9}
			\multicolumn{2}{c|}{}&&\multicolumn{2}{c|}{Training complexity}&\multicolumn{2}{c|}{Adaptation period}&\multicolumn{2}{c}{Computational costs}\\ \cline{4-9}
			\multicolumn{2}{c|}{}&&Simple&\multicolumn{1}{c|}{Difficult}&Short&\multicolumn{1}{c|}{Long}&Low&\multicolumn{1}{c}{High}\\ \hline
			\multicolumn{1}{c|}{ \multirow{6}{*}{Discriminative}}& \multicolumn{1}{c|}{Metric} &learn a similarity metric applicable cross tasks  &$\surd$&&$\surd$&  &$\surd$    &  \\ \cline{2-3}
			\multicolumn{1}{c|}{}& \multicolumn{1}{c|}{Parameter} &  Learn a task-specific predictor by a parameter generator & $\surd$  &  &$\surd$&&$\surd$&\\ \cline{2-3}
			\multicolumn{1}{c|}{}&\multicolumn{1}{c|}{Fine-tune} &  Learn a task-specific model by an optimization algorithm & $\surd$  &   &&$\surd$&&$\surd$\\  \cline{2-3}
			\multicolumn{1}{c|}{}&\multicolumn{1}{c|}{Memory} &  Cache previously seen information to enhance the features with encoders & $\surd$ &  &$\surd$&&&$\surd$\\  \cline{1-3}
			\multicolumn{2}{c|}{Generative}&  Synthesize labeled instances to enable discriminative solutions &    & $\surd$&&$\surd$&&$\surd$\\ \hline
		\end{tabular}
		\begin{tablenotes}
			\item[1]  Each technical solution is represented by its first word.
		\end{tablenotes}
	\end{threeparttable}
\end{table*}

\subsubsection{Parameter Prediction-based Methods}
Different from metric learning-based approaches focusing on learning a powerful predictor transferable cross tasks, parameter prediction-based methods target designing a unique predictor for each task. To this end, a parameter generator is devised to predict the neural weights of the prediction layer.   
In this instance, the parameters of the predictor are updated through a simple forward propagation so that fast cross-class adaptation can be achieved. However, it is difficult for the generator to estimate large-scale model parameters \cite{wu2021learning}. 

\subsubsection{Fine-tune-based Methods}
Fine-tune-based methods intend to develop an optimization algorithm, so that the segmentation model  can fast converge to model the posterior probability for a given few-shot (or zero-shot) task. Fine-tune-based methods have a powerful adaptation ability even if the unseen classes obey the different distribution from seen ones \cite{guo2020broader}.  However,  fine-tune-based methods involve some hyperparameters (i.e., the iteration steps), which are significant for model performance on target tasks.  
Moreover, due to the gradient calculation in backpropagation, a longer adaptation period is demanded to update parameters. 

\subsubsection{Memory-based Methods}
Memory-based approaches aim to cooperate with the encoder to enhance the representations of objects. Specifically, memory-based approaches take advantage of some tools, such as RNNs and memory networks,  to  store  previously seen information for assisting the representations of target classes. Thanks to the ability of these tools in capturing temporal information, memory-based approaches  play a crucial role in few/zero-shot sequential learning problems. 
Nevertheless, memory-based methods generally contribute to high computational costs.

\subsubsection{Generative Model-based Methods}
Generative model-based methods tend to build  a generator, which is learned from base classes to obtain an ability to  fit $P(x|m)$ for a given task. The generator aims to synthesize labeled instances of unseen classes (i.e., visual features), which assists in modeling the class-specific distribution and enabling the parameter adjustment on  segmentation models.    
Moreover, synthesized instances of both base and unseen classes are sometimes allowed to  train the classifier together to alleviate the prediction bias towards base classes \cite{pourpanah2020review}.  
Whereas, the distribution of generated instances fails to obey the real distribution in most cases, and the training difficulties and inference costs are intractable in few/zero-shot problems.

\subsubsection{Summary}
Apart from the distinctive pros and cons of the five technical solutions mentioned above, the comparison  of training complexity, adaptation period, and computational costs are further illustrated in Table \ref{ad}. In terms of training complexity, since the generator in generative model-based methods needs to be trained alone, the training step is more difficult than other methods. In respect of the adaptation period, owing to multiple forward and backward propagations, the time costs of fine-tune-based and generative model-based methods are higher than other solutions. Referring to computational costs, thanks to the single forward propagation and similarity calculation between a few features, metric learning-based and parameter prediction-based approaches have lower computational overhead. 
Besides, as for applicable scenarios, due to the access to the distribution of unseen classes via $D_{\rm{support}}$, few-shot circumstances incline to adopt discriminative solutions, while  zero-shot cases have a preference for generative model-based methods as  their unavailability on the distribution of target unseen classes. 
Furthermore, thanks to the high flexibility of metric learning-based methods and the excellent ability of memory-based ones in storing valuable information, metric learning-based and memory-based approaches have made outstanding contributions in both few-shot and zero-shot scenarios.

\section{Image Semantic Segmentation}\label{section5}

In this section, we review  recently proposed ISS studies that are exposed to only a few or zero annotated training samples  and group them into few-shot ISS methods and zero-shot ISS methods, according to whether the annotated samples are accessible or not.

\subsection{Few-shot Image Semantic Segmentation}
Learning a pixel-wise classifier for unseen categories from a small number of labeled samples has attracted more and more attention. In recent years, several few-shot ISS methods \cite{dong2018few,wang2019panet,zhang2019canet} have been proposed. 
Since the methods based on generative models require higher training difficulties and inference costs to generate pseudo-labeled samples, few-shot ISS generally tends to  discriminative solutions, which will be demonstrated as follows.
\subsubsection{Metric Learning-based Methods}
Metric learning plays a vital role in tackling few-shot ISS, where the approaches \cite{dong2018few,wang2019panet,zhang2019canet,rakelly2018conditional,tian2020prior,chen2021semantically,lang2022learning}  based on prototype networks \cite{snell2017prototypical} are in a dominant position. Different from the conventional learning-based approaches \cite{zhou2022rethinking,wang2021exploring,michieli2021prototype,hwang2019segsort,ke2021prototypical}, where the learned prototype of a class  is an approximate estimate of the optimal prototype, these few-shot approaches \cite{dong2018few,wang2019panet,zhang2019canet,rakelly2018conditional,tian2020prior,chen2021semantically,lang2022learning} aim to obtain  a class-specific prototype, which may not be an approximation of the optimal prototype, as long as it can provide the information of objects and enable higher similarity scores for the query features that have the same semantic classes as the prototype. 
However, it is insufficient to describe a category only by a vectorial prototype, which inspires some methods \cite{yang2021part,zhang2021rich,li2021adaptive} aiming at generating multiple prototypes for each class. In addition, directly performing element-level dense matching between support and query features \cite{zhang2019pyramid,wang2020few,liu2021harmonic} is also a feasible way to break through this obstacle. The relevant approaches are depicted as follows. 

Early approaches  \cite{dong2018few,wang2019panet,zhang2019canet} attempt to conduct feature matching with a single class descriptor.  Prototype networks were first exploited  for segmentation on unseen objects with the assistance of a few labeled samples in \cite{dong2018few}. 
Specifically,   a two-branch network was designed to estimate segmentation maps for query images. The first branch took the support set as the input and output a  global class descriptor, while the second branch leveraged the generated prototype as the guidance to tune the segmentation results of the query set.  
The work in  \cite{dong2018few} inspired the follow-up research in  \cite{wang2019panet}, where  binary segmentation was directly  completed  based on the cosine distance calculated  between the feature vectors located on  query features and the class-specific prototype \cite{wang2019panet}. To make full use of the support information,  the query samples and their predicted masks were further regarded as a new support set to guide the segmentation of the original support samples. 
Unlike the work mentioned above \cite{wang2019panet} which applies a fixed distance function, other approaches \cite{zhang2019canet,rakelly2018conditional,tian2020prior,chen2021semantically,lang2022learning} utilize a learnable neural network to predict the segmentation maps, which can be regarded as a learnable distance metric  to implicitly measure the similarity between support and query features. These approaches fuse  support cues of target classes with query features and decode the fused features to output  final segmentation results.   A common fusion strategy is to concatenate query features with the tiled  prototypes \cite{zhang2019canet,lang2022learning} or support feature maps \cite{rakelly2018conditional} along the channel dimension, where multi-scale features \cite{tian2020prior} and multi-class label information \cite{chen2021semantically} are considered to  enhance the representations of query samples. Apart from concatenating directly along the channel dimension \cite{rakelly2018conditional,zhang2019canet,chen2021semantically,tian2020prior,lang2022learning}, other methods, such as element-level addition \cite{hu2019attention},  re-weighted by attention map \cite{yang2020new}, and similarity guidance \cite{zhang2020sg,liu2022axial}, are also some feasible ways to conduct  the integration between support and query features.

Though these methods mentioned above have made undeniable contributions to few-shot ISS, some of them \cite{tian2020prior,chen2021semantically,lang2022learning}  apply the masked average pooling operation to generate a holistic descriptor for each semantic category, giving rise to some issues. 
Firstly, the insufficiency of annotated category-specific samples makes the prototype learner fail to output a  robust class representation. Secondly, due to the appearance variations between support and query samples, it is challenging to capture rich and fine-grained semantic information only by a global feature vector. To deal with this dilemma, some follow-up approaches attempt to generate multiple prototypes for each semantic category \cite{yang2021part,zhang2021rich,li2021adaptive} or conduct dense matching between support and query images \cite{zhang2019pyramid,wang2020few,liu2021harmonic}.

From the perspective of generating multiple prototypes for each class, the similarity measurement is conducted between  each generated prototype and query features. A common way is to divide the object into different parts according to a particular mechanism and generate a corresponding prototype for each part. Object semantics were decomposed to assist in the generation of multiple prototypes \cite{yang2021part}. To change the number of prototypes adaptively, similar support feature vectors with different  spatial positions were grouped  to generate a specific prototype in \cite{li2021adaptive}. In order to obtain more fine-grained feature representations, multiple part-aware prototypes  were further refined with the help of unlabeled samples in \cite{liu2020part}. In addition, 
three different descriptors were designed from multiple aspects for a specific object, which would be employed to  conduct feature matching with query features \cite{zhang2021rich}.  
Instead of obtaining  deterministic prototypes, the distribution of generated prototypes  was  estimated  to simulate the uncertainty caused by limited training images and object variations, which improved the robustness  of the segmentation model \cite{wang2021variational}. 

From the perspective of dense matching, a pyramid graph network was presented  to capture the dense correspondences between support and query features at different scales \cite{zhang2019pyramid}. A democratic attention network was  proposed  to focus more on pixels where the object was located, building a robust correspondence between support and query images \cite{wang2020few}. A harmonic feature activation strategy was  proposed, which jointly exploited exclusive support features  for pixel-level semantic matching \cite{liu2021harmonic}.  A novel cross-attention mechanism was  proposed  for aggregating more relevant pixel-wise features in support images into query ones  \cite{zhang2021few}. A bipartite graph was built 
and a graph attention mechanism as well as weight adjustment strategy were applied to promote more target-object pixels to participate in the segmentation on query images \cite{gao2022mutually}. The dense correlations of foreground and background were explored, which alleviated the information loss caused by prototype learning and dense matching of a foreground feature pair \cite{shi2022dense}.

\subsubsection{Parameter Prediction-based Methods}
In few-shot ISS, parameter prediction-based methods are frequently employed to modify the weights of  the classifier for cross-class adaptation, as demonstrated in Fig. \ref{typical}(b). By employing this, the segmentation network trained on base classes can quickly enhance the segmentation ability on unseen classes. 

A two-branch network, consisted of a conditional branch and a segmentation branch, was proposed   to tackle cross-category segmentation in \cite{shaban2017one}. The conditional branch input a class-specific image with its mask and predicted the weights of the logistic regression layer for adapting to a target object. Unlike the conditional branch, the predominant duty of the segmentation branch was to extract high-level semantic features from query images. Through the logistic regression layer with replaced parameters, the pixel-wise semantic labels could be generated from the extracted query features.  Instead of leveraging support samples merely, query images were also employed to the generation of classifier weights \cite{zhuge2021deep}. In contrast to replacing the classifier parameters directly,  the weights of the classifier were added dynamically so that the model can master both base and unseen categories in \cite{liu2020dynamic}. 

\subsubsection{Fine-tune-based Methods}
The fine-tune-based few-shot ISS aims to adopt an optimization algorithm to refine the parameters of the pre-trained segmentation network for learning unseen categories. 
The segmentation network was refined  iteratively by minimizing the error calculated from support predictions and their corresponding masks in \cite{yang2020brinet}. With the help of parameter refinement,  the performance degradation resulting from the inter-class gap between the offline and online stages was alleviated. 
An embedding network and a differentiable linear classification model were  proposed  so that the parameters of the linear classification model could be updated more efficiently while the embedding network generalizing among diverse classes in \cite{tian2020differentiable}. 
Different from the approaches \cite{yang2020brinet,tian2020differentiable} adopting episode training in the offline stage, 
 a transductive inference strategy based on standard supervised learning was resorted 
 to obtain a feature extractor on base classes \cite{boudiaf2021few}. In the inference phase,  a linear classifier was refined by means of minimizing a  loss function based on labeled support images and the statistical characteristics of unlabeled query ones. Apart from the cross-category adaptation, the shift caused by distribution diversity between training and inference data was also concerned, making it more desirable for real applications.

\subsubsection{Memory-based Methods}
In the memory-based few-shot ISS, the previously seen  information 
is reserved to help the segmentation on query samples. 
Common attributes and prior information of diverse categories with noticeable visual differences were stored into an external memory, which was manipulated to transfer labels for unseen classes \cite{wu2021learning}. 
However,  the computational loads of the model  increased with the growth of memory size \cite{wu2021learning}.  The  features of target classes with different resolutions were memorized, which reduced the computational consumption \cite{xie2021few}. The stored  features were then extracted to obtain more cross-resolution information and accurate segmentation results.

\begin{table*}[!t]
	\centering
	\begin{threeparttable}[b]
		\caption{A summary of few/zero-shot ISS methods involved in this survey.}
		\renewcommand\arraystretch{1.15}
		\label{iss}
		\begin{tabular}{c|c|cc|ccccc|c}
			\hline
			\multirow{3}{*}{\textbf{Methods}} & \multirow{3}{*}{\textbf{Years}}  & \multicolumn{2}{c|}{\textbf{Scenarios}} & \multicolumn{5}{c|}{\textbf{Technical Solutions}}   & \multirow{3}{*}{\textbf{Main contributions}}\\ \cline{3-9}
			&   & \multicolumn{1}{c|}{ \multirow{2}{*}{Few-shot}} & \multirow{2}{*}{Zero-shot} & \multicolumn{4}{c|}{Discriminative}&\multirow{2}{*}{Generative} &\\ \cline{5-8}
			&&\multicolumn{1}{c|}{}&&Metric & Parameter & Fine-tune & \multicolumn{1}{c|}{Memory}&& \\ \hline
			Shaban \emph{et al.} \cite{shaban2017one} &  2017  &   $\surd$ & &  &$\surd$ &&&& Two-branched architecture\\
			Dong \emph{et al.} \cite{dong2018few} &  2018  &  $\surd$ & & $\surd$ & &&&&Prototype learning \\
			Rakelly \emph{et al.} \cite{rakelly2018conditional}&  2018  &  $\surd$ & & $\surd$ & &&&& FCNs\\
			Wang \emph{et al.} \cite{wang2019panet}&  2019  &  $\surd$ & & $\surd$ & &&&&  Parametric classification\\
			Zhang \emph{et al.} \cite{zhang2019canet}&  2019  &  $\surd$ & & $\surd$ & &&&& Mask refinement\\
			Hu \emph{et al.} \cite{hu2019attention}&  2019  &  $\surd$ & & $\surd$ & &&&& Multi-scale feature fusion\\
			Zhang \emph{et al.} \cite{zhang2019pyramid}&  2019  &  $\surd$ & & $\surd$ & &&&&GNN, attention mechanism \\
			Tian \emph{et al.} \cite{tian2020prior}&  2020  &  $\surd$ & & $\surd$ & &&&& Prior information\\
			Yang \emph{et al.} \cite{yang2020new}&  2020  &  $\surd$ & & $\surd$ & &&& & Transformation module\\
			Zhang \emph{et al.} \cite{zhang2020sg}&  2020  &  $\surd$ & & $\surd$ & &&&& Feature representation\\
			Liu \emph{et al.} \cite{liu2020part}&  2020  &  $\surd$ & & $\surd$ & &&&&Multi-prototype representation \\
			Wang \emph{et al.} \cite{wang2020few}&  2020  &   $\surd$ & & $\surd$ & &&&&Attention mechanism\\
			Liu \emph{et al.} \cite{liu2020dynamic} &  2020  &   $\surd$ & &  &$\surd$ &&&&Attention mechanism\\
			Yang \emph{et al.} \cite{yang2020brinet}& 2020  &   $\surd$ & &  & & $\surd$ &&&Online refinement strategy\\
			Tian \emph{et al.} \cite{tian2020differentiable}&  2020  &  $\surd$ & &  & &$\surd$&&&Linear classifier\\
			Liu \emph{et al.} \cite{liu2021harmonic}&  2021  &  $\surd$ & & $\surd$ & &&&&Harmonic feature activation \\
			Li \emph{et al.} \cite{li2021adaptive}&  2021  &   $\surd$ & & $\surd$ & &&&& Multi-prototype representation\\
			Yang \emph{et al.} \cite{yang2021part}&  2021  &  $\surd$ & & $\surd$ & &&&& Multi-prototype representation\\
			Zhang \emph{et al.} \cite{zhang2021rich}&  2021  &   $\surd$ & & $\surd$ & &&&& Multi-prototype representation\\
			Wang \emph{et al.} \cite{wang2021variational}&  2021  &   $\surd$ & & $\surd$ & &&&& Probabilistic
			framework\\
			Zhang \emph{et al.} \cite{zhang2021few}&  2021  &   $\surd$ & & $\surd$ & &&&&Transformer \\
			Zhuge \emph{et al.} \cite{zhuge2021deep} &  2021  &   $\surd$ & &  &$\surd$ &&& &Feature fusion\\
			Boudiaf \emph{et al.} \cite{boudiaf2021few} &  2021  &  $\surd$ & &  & &$\surd$&&& Transductive inference\\
			Wu \emph{et al.} \cite{wu2021learning}&  2021  &  $\surd$ & &  & &&$\surd$&&Memory network \\
			Xie \emph{et al.} \cite{xie2021few} &  2021  &   $\surd$ & &  & &&$\surd$&&Memory network \\
			Chen \emph{et al.} \cite{chen2021semantically}&  2022  &  $\surd$ & & $\surd$ & &&&&Multi-class guidance\\
			Lang \emph{et al.} \cite{lang2022learning}&  2022  &  $\surd$ & & $\surd$ & &&&&Base-class prediction \\
			Shi \emph{et al.} \cite{shi2022dense}&  2022  &  $\surd$ & & $\surd$ & &&&&Feature fusion\\
			Liu \emph{et al.} \cite{liu2022axial} & 2022&  $\surd$ & & $\surd$ & &&&&Weight-sparsification\\
			Gao \emph{et al.} \cite{gao2022mutually} &  2022  &  $\surd$ & & $\surd$ & &&&& Attention mechanism\\ \cline{1-10}
			Xian \emph{et al.} \cite{xian2019semantic} &  2019  &   &$\surd$ & $\surd$ & &&&& Projection function\\
			Kato \emph{et al.} \cite{kato2019zero}&  2019  &   &$\surd$ & $\surd$ & &&&&Projection function \\
			Bucher \emph{et al.} \cite{bucher2019zero} & 2019 & &$\surd$ & & &&&$\surd$&Zero-shot framework \\
			Tian \emph{et al.} \cite{tian2020cap2seg} &  2020  &   &$\surd$ & $\surd$ & &&&&Image captions\\
			Gu \emph{et al.} \cite{gu2020context}& 2020 &   &$\surd$ & & &&&$\surd$ &Feature generator \\
			Li \emph{et al.} \cite{li2020consistent}& 2020 &  &$\surd$ & & &&&$\surd$&Feature generator  \\
			Hu \emph{et al.} \cite{hu2020uncertainty} &  2020  &  &$\surd$ & $\surd$ & &&& &Uncertainty estimation\\
			Lu \emph{et al.} \cite{lu2021feature} & 2021  &   &$\surd$ & $\surd$ & &&&&Feature enhancement\\
			Baek \emph{et al.} \cite{baek2021exploiting} & 2021 &   &$\surd$ & $\surd$ & &&&& Loss function\\
			Gu \emph{et al.} \cite{gu2020pixel}& 2022 &   &$\surd$ & & &&&$\surd$ &Feature generator\\\hline
		\end{tabular}
		\begin{tablenotes}
			\item[1]  Each technical solution is represented by its first word.
		\end{tablenotes}
	\end{threeparttable}
\end{table*}

\subsection{Zero-shot Image Semantic Segmentation}

Zero-shot ISS focuses on performing segmentation for zero-label categories unavailable in the training process. Attributable to the absence of supervision signals on class-specific visual samples, it is tricky for zero-shot ISS to modify the model weights directly. Consequently, flexible metric learning-based settlements and generative model-based methods capable of generating pseudo labeled data are adopted by zero-shot ISS  for learning new categories.

\subsubsection{Metric Learning-based Methods}
In zero-shot ISS, the features encoded from query samples are in a latent visual space while the word embeddings are in a semantic space.  Due to the inconsistency between the two embedding spaces, it is unreasonable to carry out feature matching directly. Therefore, the metric learning-based approaches endeavor to map the visual and/or semantic features into a shared space, which can be the semantic space \cite{xian2019semantic,tian2020cap2seg,lu2021feature}, the visual space \cite{kato2019zero} and the shared latent space \cite{baek2021exploiting},  so that  the distance measurement can be conducted.

\begin{figure}[!t]
	\centering
	\includegraphics[width=0.9\linewidth]{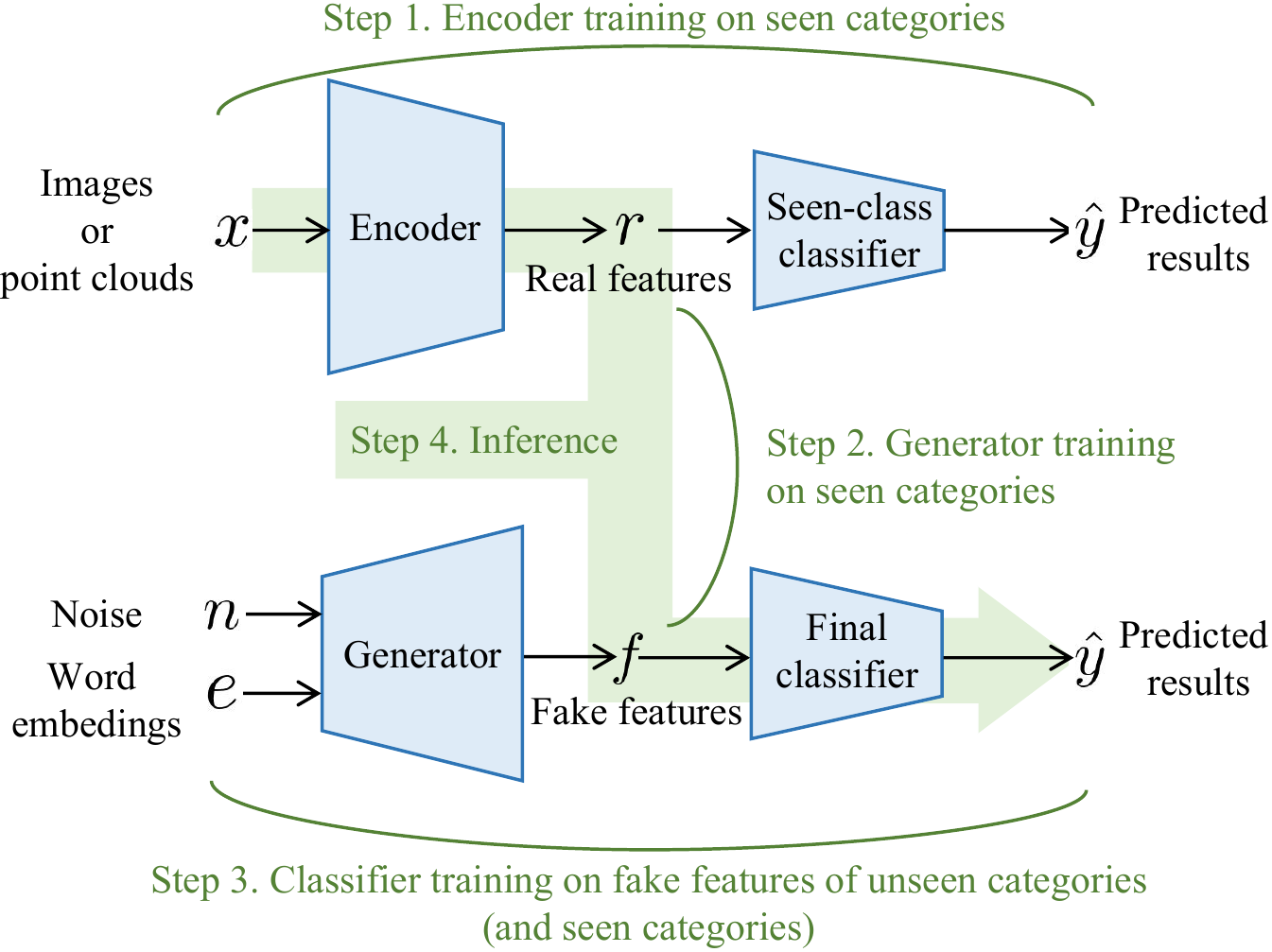}
	\caption{Implementation process of zero-shot ISS and zero-shot 3DS approaches based on generative model (adapted from \cite{bucher2019zero,michele2021generative})}
	\label{generative}
\end{figure}

The majority of these approaches \cite{xian2019semantic,tian2020cap2seg,lu2021feature}  take the semantic space as the common embedding space. Thus, the function projecting visual features into the semantic space should be devised in advance.  Zero-annotation ISS on novel categories was conducted  by two steps  \cite{xian2019semantic}. Firstly, a visual-semantic embedding module was devised to project the class-specific visual information into the semantic space, where every mapped embedding vector could be regarded as the  representation of a particular pixel of the query sample. Secondly, a mask was predicted based on the similarity between the semantic and pixel embeddings. Despite that achievements have been made in  \cite{xian2019semantic}, the prediction bias towards observed classes arises when transferring the knowledge from the observed classes to unseen categories. 
To cope with this issue, the studies in  \cite{tian2020cap2seg,lu2021feature} endeavor  to train the model on observed categories together with the information from unseen ones. Image captions rather than word embeddings were adopted to dig out supervision signals for unseen classes, where the position cues of unseen classes could be provided by base ones \cite{tian2020cap2seg}. 
Target-class semantic embeddings were utilized  in training to reduce the prediction bias, which aimed to learn the shareable information concealed in the source and target semantic embeddings  \cite{lu2021feature}.  Specifically, a saliency detection strategy was proposed to effectively distinguish the area where the source objects were located  so that the pixel-level label assignment for both source and target classes could be conducted in their respective areas. 
 Nevertheless, these methods \cite{xian2019semantic,lu2021feature} do not pay attention to the unfavorable impact resulting from noisy and outlying samples, which contributes to biased estimation of target classes \cite{hu2020uncertainty}. Based on this, Bayesian uncertainty estimation \cite{kendall2017uncertainties} was introduced to aggregate more discriminative samples on seen classes to carry out pixel-wise label prediction \cite{hu2020uncertainty}. However, extra parameters were involved in estimating the uncertainty of input images.

There are also some approaches \cite{kato2019zero,baek2021exploiting} conducting zero-shot ISS in other embedding spaces. Distinct from classifying pixels in the semantic space  \cite{xian2019semantic,tian2020cap2seg,lu2021feature},   the segmentation of unseen classes was also allowed to be conducted in the visual space \cite{kato2019zero}. The potential distribution of  semantic embeddings was constructed, where the embeddings  were then decoded with visual features  to measure the pair-wise similarity. 
A joint embedding space was opted to train a visual encoder and a semantic encoder in  \cite{baek2021exploiting}. Moreover,  two complementary loss functions were presented to learn more representative embeddings and a decision boundary modification scheme was proposed for the prediction bias.

\subsubsection{Generative Model-based Methods}

It can be seen in Fig. \ref{typical}(e) that the generative model-based methods strive to estimate labeled instances  to enable the weight modification on the classifier.  As one of the representative methods, the encoder was trained  firstly on seen categories to estimate real visual features in \cite{bucher2019zero}. The encoded  features  and  semantic embeddings of seen classes were  then exploited  to learn a generator, which could take  semantic embeddings as input and output the fake visual features for unseen categories. The generated features of never-seen categories (and real features of seen categories) were finally applied to refine the classifier weights at inference time, as illustrated in Fig. \ref{generative}. 
The research in  \cite{bucher2019zero} spawns some variants \cite{gu2020context,li2020consistent,gu2020pixel}, where contextual information \cite{gu2020context,gu2020pixel} and inter-class structural relationship  \cite{li2020consistent} were considered to synthesize higher-quality visual features.

\begin{figure}[!t]
	\centering
	\includegraphics[width=0.95\linewidth]{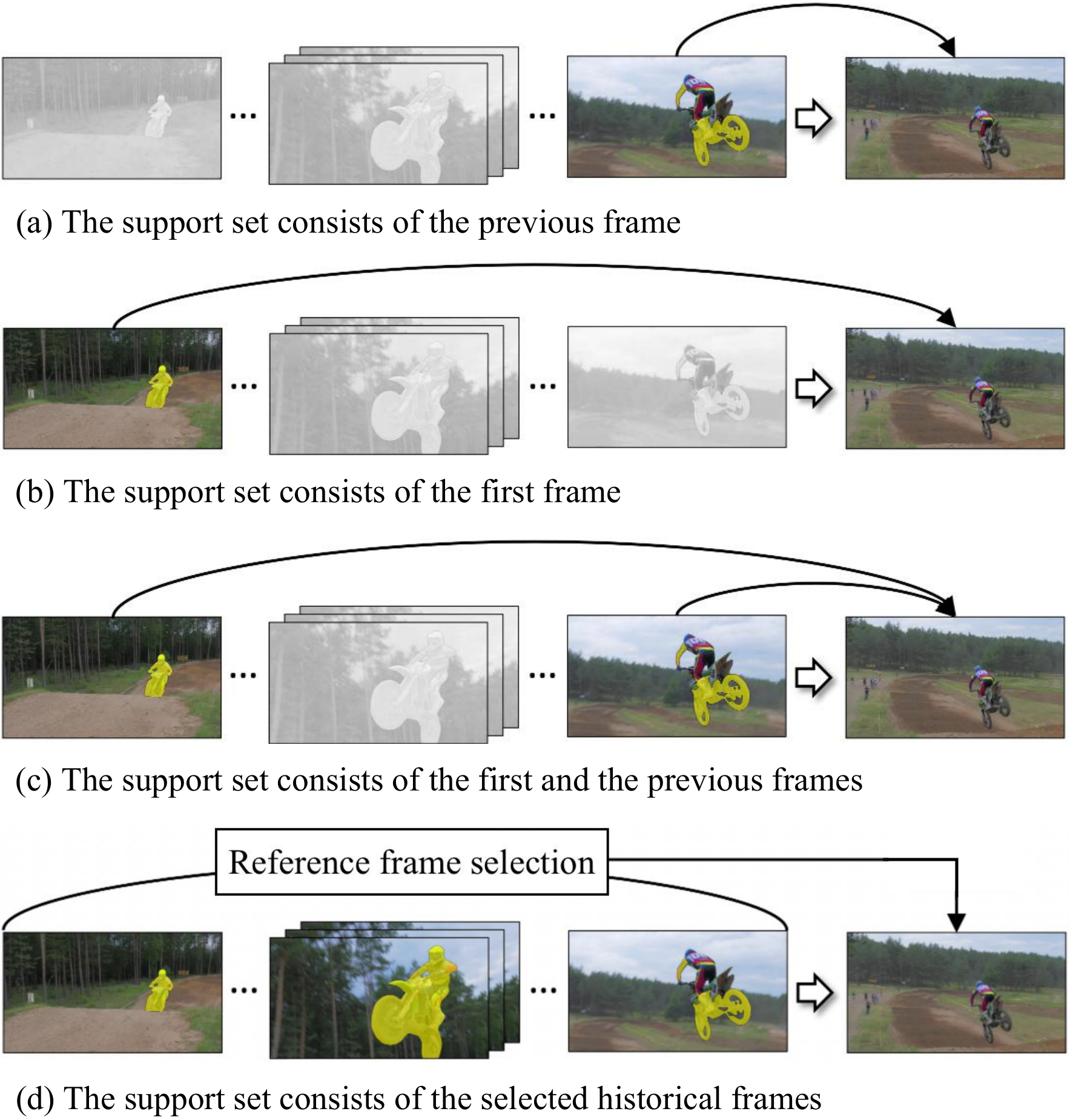}
	\caption{The settings of the support set in few-shot VOS (adapted from \cite{oh2019video}), where the mask of the first frame is the ground truth and the masks of intermediate and previous ones are self-estimated masks.  These constructions are also commonly-used in zero-shot VOS, where the mask of the first frame is unavailable.}
\label{selection}
\end{figure}

\subsection{Summary}
 To elaborate this section intelligibly, we further describe  the involved few/zero-shot ISS approaches  in Table \ref{iss}.  It can be seen that the methods based on metric learning occupy a crucial position in few-shot ISS, where the diversity of feature representation, distance fuctions and feature matching strategies give researchers inexhaustible motivation for innovations. 
Due to the characteristics of zero annotation in zero-shot ISS, it is challenging to directly leverage semantic embeddings to optimize visual segmentation models. Therefore, easy-to-implement metric learning-based approaches are preferred by zero-shot ISS. Moreover, adopting generative model-based approaches to convert zero-shot ISS into few-shot problems is also popular in breakthrough these limitations.

\section{Video Object Segmentation}\label{section6}
Few/zero-shot VOS is the extension of few/zero-shot  ISS in temporal dimension. When the objects in a  task have the same identities and support and query images are continuous  on time stamps,  a few/zero-shot  ISS problem becomes a few/zero-shot  VOS challenge.  Compared with few/zero-shot  ISS, few/zero-shot  VOS has no requirement on manual construction of few/zero-shot tasks, where a video of an unseen class can be naturally regarded as a target task.  The common VOS settings of the support set are demonstrated in Fig. \ref{selection}. The previous frame with the predicted mask and/or the first frame with given annotations can be adopted to guide the segmentation of the current frame, and the intermediate frames spanning from the first to previous frames are also allowed to be leveraged. Moreover, compared with ISS, few/zero-shot  VOS burdens fewer appearance variations and enjoys richer temporal information, where appearance and temporal cues in support samples can be applied to enhance the representation of query samples, leading to memory-based methods preferred in this case.

\subsection{Few-shot Video Object Segmentation}
Due to the inter-frame similarity of the object appearance in a video, it is reasonable to adopt metric learning-based approaches to separate the area similar to the given object from the current frame. In addition, for a given video, the previously seen historical frames play a considerable role in enhancing the representations of objects in the current frame, leading to the prosperity of memory-based approaches. Different from few-shot ISS, few-shot VOS has higher requirements on the  speed of model adaptation, which is desired to be larger than the frame rate. However, owing to the problem of time-consuming, fine-tune-based approaches are facing a decline. In this section, we will describe these solutions separately.

\subsubsection{Metric Learning-based Methods}
Compared with few-shot ISS, only the first frame is labeled in few-shot VOS. It is challenging to construct a reliable class descriptor merely based on a single frame. Thus, few-shot VOS usually carries out dense matching between the features generated by current and reference frames. 
 In this section, we categorize and review each method according to the different selection of reference frames.

The  first  frame \cite{wang2019fast,yin2021directional} or the previous frame \cite{motionguide} (or both \cite{voigtlaender2019feelvos,yang2020collaborative,zhu2021separable}) is to be given preference by earlier studies for comprising support sets, as depicted in Fig. \ref{selection}(a)-(c). 
Object tracking was combined with VOS  to estimate masks as well as rotated bounding boxes simultaneously, with the assistance of the pairwise similarity between the features mapped from the current frame and the initial template \cite{wang2019fast}. Taking the low efficiency caused by pixel-wise dense matching and the curse of dimension in the Euclidean space \cite{jain2000statistical} into account,  a hypersphere embedding space was built, where  the cosine distance was replaced with a  convolution layer  to accelerate the similarity measurement \cite{yin2021directional}. As an alternative to the initial frame, the previous frame adjacent to the current frame was employed to provide appearance information of the target object in \cite{motionguide}. 
By measuring the distance between the features generated from the previous and current frames, a coarse segmentation map of the current frame could be obtained, which would be further refined for more accurate prediction. 
For obtaining global and local correlations, a novel pixel-level matching mechanism was presented, where the current frame was employed to calculate correlations with both first and historcal frames in \cite{voigtlaender2019feelvos}. 
To alleviate mismatching on similar objects in the background, foreground and background areas were addressed  identically in \cite{yang2020collaborative}. Taking the appearance and structure information on the target objects into account, a structure modeling branch was constructed  to encode the information of the complete object and its components in \cite{zhu2021separable}.

\begin{figure*}[!t]
	\centering
	\includegraphics[width=\linewidth]{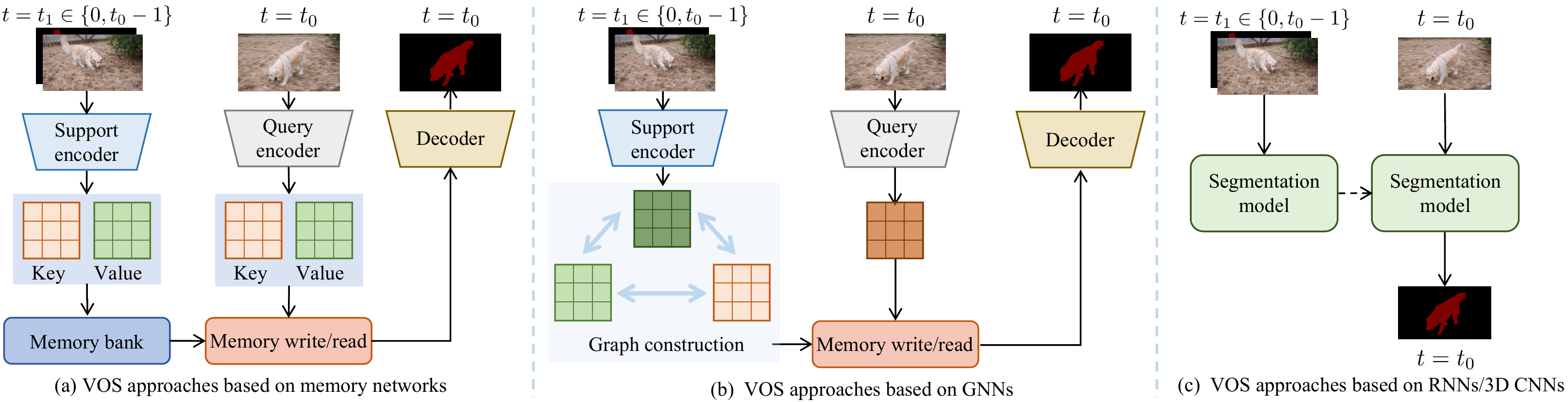}
	\caption{ Data flows in memory-based VOS methods, in which the masks of historical frames are only leveraged in few-shot scenarios. Memory networks and GNNs are adopted to cache the valuable information explicitly, while RNNs and 3D CNNs, which are leveraged as some subparts of the segmentation model, are employ to store the knowledge implicitly. The squares with different colors represent diverse features.}
	\label{memory_few}
\end{figure*}

Nevertheless, these approaches  \cite{wang2019fast,yin2021directional,motionguide,voigtlaender2019feelvos,yang2020collaborative,zhu2021separable} 
only leverage the initial frame and/or the adjacent frame, making it tricky to deal with the challenge of object deformation, occlusion, and model shift. Therefore, other metric learning-based approaches \cite{zhang2020transductive,liu2020guided,hong2021adaptive} aim to explore a reasonable intermediate frame sampling mechanism, as illustrated in Fig. \ref{selection}(d),  to capture richer temporal information and tackle the shift caused by sparse matching. To explain the appearance variations of the target objects, more recent and less long-distance frames were selected in \cite{zhang2020transductive}. An observed video was clipped  into different snippets, from each of which  a frame was selected  to build a support set together with the labeled first frame, and the final segmentation map was obtained by averaging the predictions conducted on the target frame and each support frame in  \cite{liu2020guided}. Different from the manually designed sampling strategy \cite{zhang2020transductive,liu2020guided}, an adaptive selection mechanism was presented based on the similarity in object appearance and the accuracy on predicted masks in \cite{hong2021adaptive}. 
 Despite the utilization of intermediate frames can bring about performance gains, redundant computational consumption exists when the appearance variations of target objects are slight cross different frames. On that account, a new segmentation network was proposed  to dynamically adjust the processing strategy according to the appearance changes across frames, which reduced the unnecessary computational occupation \cite{park2021learning}.

\subsubsection{Fine-tune-based Methods}
The fine-tune-based methods endeavor to optimize the model for each video sequence based on the first labeled frame (and the subsequent frames with their self-predicted masks), where over-fitting risks, time-consuming problems and handcrafted hyperparameters hinder their development. 
To alleviate over-fitting and generalize to new video objects more flexibly,  a novel architecture consisting of a segmentation sub-network and a lightweight appearance sub-network was proposed, where only the appearance one was updated online  \cite{robinson2020learning}. 
In response to the challenges of time-consuming and handcrafted hyperparameters,  optimization-based meta-learning \cite{finn2017model} was  integrated into online adaptation for efficient VOS  \cite{xiao2019online}. In the offline stage, the common knowledge between different objects was mined by training on multiple similar segmentation tasks, and  the optimal model initialization as well as  parameter-level learning rates were provided. In the online adaptation, the first and previous frames were leveraged to fine-tuned the initialized segmentation model with provided learning rates, reaping the  characteristics of unseen objects and adapting to annoying appearance variations. 
Whereas, providing  learning rates for each parameter places restrictions on large-scale segmentation networks. To address this issue,  neuron-level learning rates were explored, which considerably eased the requirements on the number of  learning rates in  \cite{meinhardt2020make}. Moreover,  VOS could be decomposed  into object detection and local-mask prediction, so that the fine-tuning operation could be carried out with the assistance of bounding box propagations. Despite higher efficiency can be reached,  these methods \cite{xiao2019online,meinhardt2020make} conducted optimization with fixed learning rates,  making them challenging to adaptively deal with  target objects with diverse appearances \cite{xu2021meta}. To handle this,   an evaluation criterion was proposed  for the meta learner and  the learning rates were automatically modified for the upcoming frames  \cite{xu2021meta}.

\subsubsection{Memory-based Methods}
To cache previously seen information, memory networks \cite{oh2019video,oh2020space,seong2020kernelized}, GNNs \cite{zhang2020dual} and RNNs \cite{lyu2019lip,zhang2020dual}  are leveraged in memory-based methods, as shown in Fig. \ref{memory_few}. 

As one of representative approaches,  a space-time memory network was devised, which could store helpful information from observed frames  \cite{oh2019video,oh2020space}. 
Peculiarly, when the estimation was performed on the current frame, a pixel-wise matching on key maps was conducted to read out the beneficial value maps for enhancing the representations of the current frame. Furthermore, to attenuate the adverse effects resulting from the appearance deformation, the features stored in the memory were updated dynamically with the segmentation of video frames, as demonstrated in Fig. \ref{memory_few}(a). 
However, these methods \cite{oh2019video,oh2020space} conduct global-to-global dense matching strategies  between query and memory features, leading to erroneous segmentation on background objects that are similar to the target ones. To cope with the mismatching, some studies  \cite{seong2020kernelized,seong2022video,xie2021efficient,seong2021hierarchical,li2020fast,liang2020video,cheng2021rethinking} attempt to build non-global relationships between  features.  A memory network based on the Gaussian kernel was  developed to compel features   focusing on a single object, which reduced the occurrence of mismatching \cite{seong2020kernelized,seong2022video}. 
A local-to-local matching between the regions of memory and query frames was performed  \cite{xie2021efficient}, which mitigated the mismatching on similar objects and brought about lower computational consumption than global-to-global methods \cite{oh2019video,oh2020space}. Whereas, these methods \cite{oh2019video,oh2020space,xie2021efficient} only perform matching at coarse scales, which makes them challenging to capture fine-grained information \cite{seong2021hierarchical}. Considering this,  a novel reading mechanism was proposed  to set up multi-scale correspondences, where  coarse correlations of dense matching were employed  to guide sparse matching at finer scales \cite{seong2021hierarchical}. 

Nevertheless, these methods \cite{oh2019video,oh2020space,seong2021hierarchical} only conduct memorization on the latest information without abandoning the antiquated or useless one, resulting in the increasing computational costs and memory occupation with the progress of the  segmentation. Consequently, a considerable number of approaches \cite{li2020fast,liang2020video,cheng2021rethinking} have been presented to manage the memory features 
efficiently. A global representation with a fixed size was learned  from all support frames, which decoupled the computational consumption with the length of the given video sequence  \cite{li2020fast}.  A feature bank was embedded   into the memory network to 
aggregate useful features and forget worthless ones dynamically \cite{liang2020video}. Without storing the key and value maps 
for each instance individually, the correlation between query and memory frames was delved to retrieve the value map encoded from multiple instances, where a voting mechanism was employed for feature aggregation \cite{cheng2021rethinking}. Aside from CNN-based memory networks utilized by these mentioned approaches \cite{oh2019video,oh2020space,seong2020kernelized,seong2022video,xie2021efficient,seong2021hierarchical,li2020fast,liang2020video,cheng2021rethinking}, graph-based memory networks are also employed \cite{lu2020video}. In  \cite{lu2020video}, a novel graph memory network was designed to store the information of the support frames with their masks. To quickly adapt to the visual variations on different timestamps, a learnable controller was added to update features and store more abundant information dynamically  on the premise of maintaining a fixed memory size.

Furthermore, GNNs and RNNs also play a crucial role in few-shot VOS, as shown in Fig. \ref{memory_few}(b) and Fig. \ref{memory_few}(c).  GNNs and GRUs were  jointly adopted  to model short-term and long-term temporal information, respectively, where the short-term information was cached in nodes of GNNs while the long-term one is stored in the GRU modules  \cite{zhang2020dual}.  The GRU modules were also embeded into the segmentation model  to propagate features for proposal generation \cite{lyu2019lip}.

\begin{table*}[!t]
	\centering
	\begin{threeparttable}
		\caption{A summary of few/zero-shot VOS methods. }
		\renewcommand\arraystretch{1.15}
		\label{video}
		\begin{tabular}{c|c|cc|ccc|c|c}
			\hline
			\multirow{2}{*}{\textbf{Methods}} & \multirow{2}{*}{\textbf{Years}}  & \multicolumn{2}{c|}{\textbf{Scenarios}} & \multicolumn{3}{c|}{\textbf{Technical Solutions}} & \multirow{2}{*}{\textbf{Flow}}& \multirow{2}{*}{\textbf{Main contributions}} \\
			&   &  Few-shot & Zero-shot & Metric & Fine-tune & Memory  &&   \\ \hline
			Wang \emph{et al.} \cite{wang2019fast}&  2019  &  $\surd$ & & $\surd$  &&&&Multi-task \\
			Voigtlaender \emph{et al.} \cite{voigtlaender2019feelvos}&  2019  &  $\surd$ & & $\surd$ &&& &Feature embedding, matching mechanism\\
			Khoreva \emph{et al.} \cite{khoreva2019lucid}&  2019  & $\surd$ & & & $\surd$ &&$\surd$&Data augmentation \\
			Xiao \emph{et al.} \cite{xiao2019online}&  2019   & $\surd$ & & & $\surd$ &&&Meta learner \\
			Oh \emph{et al.} \cite{oh2019video}&  2019  & $\surd$ & & &  &$\surd$&&Memory network,  matching mechanism\\
			Lyu \emph{et al.} \cite{lyu2019lip}&  2019  & $\surd$ & & &  &$\surd$&&Conv-GRU module, multi-task  \\
			Hu \emph{et al.} \cite{motionguide} &  2020  & $\surd$ & & $\surd$  &&&$\surd$&Refinement network, attention mechanism  \\
			Yang \emph{et al.} \cite{yang2020collaborative}&  2020  & $\surd$ & & $\surd$  &&&& Feature embedding\\ 
			Zhang \emph{et al.} \cite{zhang2020transductive} &  2020  & $\surd$ & & $\surd$  &&&&Transductive inference \\
			Liu \emph{et al.} \cite{liu2020guided}&  2020  & $\surd$ & & $\surd$  &&& &Reference frame selection\\
			Robinson \emph{et al.} \cite{robinson2020learning}&  2020   & $\surd$ & &&  $\surd$ &&& Optimization technique\\
			Meinhardt \emph{et al.} \cite{meinhardt2020make}&  2020 & $\surd$ & &&  $\surd$ &&& Optimization technique\\
			Seong \emph{et al.} \cite{seong2020kernelized}&  2020   & $\surd$ & &&  &$\surd$&& Memory network\\
			Li \emph{et al.} \cite{li2020fast}&  2020   & $\surd$ & &&   &$\surd$&& Feature embedding\\
			Liang \emph{et al.} \cite{liang2020video}&  2020  & $\surd$ & &&   &$\surd$&&Memory network,  loss function \\
			Lu \emph{et al.} \cite{lu2020video}&  2020   & $\surd$ & &&   &$\surd$&&Memory network \\
			Zhang \emph{et al.} \cite{zhang2020dual}&  2020   & $\surd$ & &&   &$\surd$&& Memory network\\
			Yin \emph{et al.} \cite{yin2021directional}&  2021   & $\surd$ & & $\surd$  &&& &Matching mechanism, feature embedding\\
			Park \emph{et al.} \cite{park2021learning}&  2021   & $\surd$ & & $\surd$  &&& & Dynamic network\\
			Xu \emph{et al.} \cite{xu2021meta}&  2021   & $\surd$ & & & $\surd$ &&$\surd$ &Optimization technique, loss function\\
			Xie \emph{et al.} \cite{xie2021efficient}&  2021  & $\surd$ & &&   &$\surd$&$\surd$& Memory network\\
			Seong \emph{et al.} \cite{seong2021hierarchical}&  2021  & $\surd$ & &&   &$\surd$&& Memory network,  matching mechanism\\
			Cheng \emph{et al.} \cite{cheng2021rethinking}&  2021   & $\surd$ & & &  &$\surd$&&Memory matching mechanism  \\
			Zhu \emph{et al.} \cite{zhu2021separable}&  2022  & $\surd$ & & $\surd$  &&&& Structure modeling\\
			Hong \emph{et al.} \cite{hong2021adaptive}& 2022  & $\surd$ & &$\surd$ &  &&  &Reference frame selection\\
			Oh \emph{et al.} \cite{oh2020space}&  2022  & $\surd$ & &&   &$\surd$&&Memory network \\
			Seong \emph{et al.} \cite{seong2022video}&  2022   & $\surd$ & & & &$\surd$&&Memory network \\ \cline{1-9} 
			Yang \emph{et al.} \cite{yang2019anchor}&  2019  &  &$\surd$ &$\surd$& &&   &Aggregation
			technique\\
			Zhuo \emph{et al.} \cite{zhuo2019unsupervised}&  2019   &  &$\surd$ &$\surd$&  && $\surd$& Feature fusion\\
			Wang \emph{et al.} \cite{wang2019zero}&  2019  &   &$\surd$ &&   &$\surd$&&GNN \\
			Ventura \emph{et al.} \cite{ventura2019rvos}&  2019   &  &$\surd$ &&   &$\surd$&& ConvLSTM decoder\\
			Tokmakov \emph{et al.} \cite{tokmakov2019learning}&  2019   &  &$\surd$ &&   &$\surd$&$\surd$&  ConvGRU module\\
			Gu \emph{et al.} \cite{gu2020pyramid}&  2020  &  &$\surd$ &$\surd$&  &&  &Self-attention mechanism\\
			Zhou \emph{et al.} \cite{zhou2020motion}&  2020   &  &$\surd$ &$\surd$&   && $\surd$&Attention mechanism\\
			Seo \emph{et al.} \cite{seo2020urvos}&  2020   &  &$\surd$ &$\surd$&   &&  &Attention mechanism\\
			Bellver \emph{et al.} \cite{bellver2020refvos}&  2020   &  &$\surd$ &$\surd$&   &&  &Non-trivial referring expressions\\
			Mahadevan \emph{et al.} \cite{mahadevan2020making}&  2020   &  &$\surd$ & &  &$\surd$&& 3D CNN decoder \\
			Liu \emph{et al.} \cite{liu2021f2net}&  2021  &  &$\surd$ &$\surd$&   &&&Feature fusion, matching mechanism \\
			Zhou \emph{et al.} \cite{zhou2021flow}&  2021  &  &$\surd$ &$\surd$&   &&$\surd$ & Feature refinement\\
			Yang \emph{et al.} \cite{yang2021learning}&  2021 &  &$\surd$ &$\surd$&   && $\surd$&Attention mechanism, feature fusion\\
			Ji \emph{et al.} \cite{ji2021full}&  2021   &  &$\surd$ &$\surd$&   && $\surd$& Attention mechanism, feature refinement\\
			Zhao \emph{et al.} \cite{zhao2021multi}&  2021  &  &$\surd$ &$\surd$&   && $\surd$&Multi-source fusion\\
			Yang \emph{et al.} \cite{yang2021hierarchical}&  2021  &  &$\surd$ &$\surd$&   && $\surd$&Feature fusion \\
			Lu \emph{et al.} \cite{lu2020zero}&  2022  &  &$\surd$ &$\surd$&   &&&  Siamese network, attention mechanism\\
			Botach \emph{et al.} \cite{botach2022end}&  2022  &  &$\surd$ &$\surd$&   &&&  Transformer framework\\
			Li  \emph{et al.} \cite{li2022you}&  2022  &  &$\surd$ &&$\surd$   &&&  Feature fusion, optimization technique\\ \hline 
		\end{tabular}
		\begin{tablenotes}
			\item[1] Each method is represented by its first word.
			\item[2] ``Flow" represents the methods involving optical flows.
		\end{tablenotes}
	\end{threeparttable}
\end{table*}

\subsection{Zero-shot Video Object Segmentation}
Zero-shot VOS targets separating primary moving objects without labeled samples available. As seen in Fig. \ref{support}(d), 
 distinct from zero-shot ISS and zero-shot 3DS extracting supervision signals from semantic embeddings, 
the auxiliary information in zero-shot VOS is allowed to be appearance cues (i.e., unlabeled historical frames),  motion cues (i.e., optical flow), or semantic cues (i.e., language descriptions).  
The identity of the object in zero-shot VOS, whose appearance  has strong correlations between adjacent frames, is the same among diverse frames. Therefore, it is reasonable and intuitive to  adopt  memory-based methods instead of more complex generative model-based approaches for this task, leading to some principal solutions in zero-shot VOS: metric learning-based, fine-tune-based and  memory-based settlements.

\subsubsection{Metric Learning-based Methods}
Due to the differences in auxiliary samples, the pairwise similarity can be calculated between appearance features of  historical and current frames, motion and appearance features of the current frame, and semantic features of language expressions and appearance features of frames.

From the standpoint of the similarity captured from inter-frame appearance embeddings, the dominant challenges lie in  reference frame selection and feature matching mechanisms. A feasible way is to conduct pixel-level non-local matching between initial and current frames, where inter-frame information \cite{yang2019anchor} and intra-frame information \cite{liu2021f2net} are leveraged. However, non-local matching may lead to unexpected computational expenditure. Targeting at this issue, 
 local matching  was carried out only on object locations \cite{gu2020pyramid}.   To make full utilization of the global information of a given video,  the correlation between two order-independent frames was grasped in \cite{lu2020zero}, which helped to determine the area where required to be segmented.

When it comes to the similarity calculated by the current frame and the optical flow, a well-designed fusion strategy is contrived to implicitly align motion features 
to appearance ones. 
Given a current frame, a pixel-level interaction between the prominent motion map and object proposals was  performed, which eliminated the obstacles derived from moving background and unchanging objects \cite{zhuo2019unsupervised}. 
To encourage the appearance features generated by distinct convolution stages, a multi-level interaction operation was proposed  to interplay with the motion ones, so that the representations of objects could be  decoded into more accurate segmentation masks \cite{zhou2020motion}. 
To suppress the misleading knowledge caused by noisy optical flows,  the methods of  completing the discontinuous edges of optical flows \cite{zhou2021flow}, dynamically adjusting the effects of motion and appearance cues on spatio-temporal representations \cite{yang2021learning}, and promoting consistent features and suppressing incompatible ones  \cite{ji2021full} were presented. Moreover, depth maps and static saliency were integrated, where the representations of foreground objects were enlarged and purified in  \cite{zhao2021multi}. 

The similarity also can be calculated between  semantic features of language expressions and appearance features of frames. Two attention modules were carefully designed to encourage temporal consistency and  prevent model shift in  \cite{seo2020urvos}. Trivial and non-trivial linguistic phrases were encoded into language features to identify referents more accurately in \cite{bellver2020refvos}.   To encode semantic and spatial information of objects at multiple levels, visual features of different scales interacted with linguistic features in the encoder \cite{yang2021hierarchical}.   A multimodal transformer was presented in \cite{botach2022end} to conduct cross-modal interaction between semantic and appearance features.

\subsubsection{Fine-tune-based Methods}
Since language features generally carry unrelated information of objects, fusing language features with visual ones may lead to inferior segmentation accuracy. To address this issue, a learning-to-learn pattern was proposed, where target-specific cues can be obtained by fine-tuning the parameters of the transfer function \cite{li2022you}.

\subsubsection{Memory-based Methods}
The memory tools shown in Fig. \ref{memory_few} are also applied to  zero-shot scenarios. 
RNNs were employed to perpetrate implicit memory  in  both spatial and temporal dimensions \cite{ventura2019rvos}. In the spatial dimension, RNNs were exploited to explore other objects at different locations within the given frame. In the temporal dimension, the role of RNNs was to maintain the relationships across different frames. Assisted by these settings, the representations of multiple instances could be enhanced simultaneously by only a single forward propagation. Moreover, GRUs \cite{tokmakov2019learning} and 3D CNNs \cite{mahadevan2020making} were embedded into the architecture to extract spatio-temporal information from video sequences. 
A fully connected graph was established  to mine the interrelationship between any two frames in  \cite{wang2019zero}, where video frames were encoded as nodes and their correlations were modeled as edges. When separating the object of interest frame by frame, the information stored in the graph was dynamically updated to obtain comprehensive knowledge and precise segmentation masks.

\subsection{Summary}
In the discussions of this section, it can be found that compared with ISS, few/zero-shot solutions face a great deal of  new challenges and opportunities when addressing VOS problems. The extreme inadequacy of labeled data is the first challenge, which leads to prototype networks that thrive in ISS are not suitable for VOS. Secondly, the utilization of intermediate frames may cause cumulative errors on the current frame, where a  selection mechanism to pick up high-quality reference frames needs to cooperate with few/zero-shot solutions. Thirdly,  the requirements on the real-time performance of VOS are much higher than that of ISS, which requires additional consideration when designing  frameworks. Concerning opportunities, it is worth mentioning that VOS  can access  richer information, such as the correlations among support frames and query ones, which leads to generous variants of technical solutions. 
In addition, due to the similarity among ISS and VOS tasks, some ideas proposed in ISS can be transferred into few/zero-shot VOS scenarios. Table \ref{video} summarizes the few/zero-shot VOS methods introduced in this paper.

\section{3D Segmentation}\label{section7}
Few/zero-shot 3DS can alleviate the requirements for expensive annotations of unseen classes and address cross-class adaptation as ISS and VOS problems. However, since the data is disordered and nonstructural 3D samples rather than regular RGB images, as can be seen from  Fig. \ref{support}(e) and Fig. \ref{support}(f), few/zero-shot 3DS is more challenging than 2D cases, where architectures designed for 2D samples generally fail to be employed into 3D circumstances. 
Consequently, the existing few/zero-shot 3DS  methods draw lessens from the 2D approaches and develop applicable  protocols for 3D data structures. In this section, 
we laconically review the existing few/zero-shot 3DS  approaches.

\begin{table*}[!t]
	\centering
	\begin{threeparttable}
		\caption{A summary of few/zero-shot 3DS methods. }
		\renewcommand\arraystretch{1.15}
		\label{3d}
		\begin{tabular}{c|c|cc|cccc|c|c}
			\hline
			\multirow{3}{*}{\textbf{Methods}} & \multirow{3}{*}{\textbf{Years}}  & \multicolumn{2}{c|}{\textbf{Scenarios}} & \multicolumn{4}{c|}{\textbf{Technical Solutions}}& \multirow{3}{*}{\textbf{Point} } &\multirow{3}{*}{\textbf{Main contributions} }  \\ \cline{3-8}
			&   & \multicolumn{1}{c|}{\multirow{2}{*}{Few-shot}} & \multicolumn{1}{c|}{\multirow{2}{*}{Zero-shot}} & \multicolumn{3}{c|}{Discriminative}& \multirow{2}{*}{Generative}   &&  \\ \cline{5-7}
			&&\multicolumn{1}{c|}{}&\multicolumn{1}{c|}{}&Metric & Parameter & \multicolumn{1}{c|}{Fine-tune}& & \\  \hline
			Sharma \emph{et al.} \cite{sharma2019learning}&  2019   & $\surd$ & &  & &$\surd$&& $\surd$&Embedding network\\
			Chen \emph{et al.} \cite{chen2019bae}&  2019   & $\surd$ & &  & &$\surd$&&&Autoencoder\\
			Sharma \emph{et al.} \cite{sharma2020self}&  2020  & $\surd$ & &  & &$\surd$&&$\surd$&Self-supervised pre-training\\
			Chen \emph{et al.} \cite{chen2020compositional}&  2020   & $\surd$ & & $\surd$ & &&&$\surd$&  Prototype network\\
			Yuan \emph{et al.} \cite{yuan2020ross} &  2020  & $\surd$ & & $\surd$ & &&&&Descriptor generator \\
			Wang \emph{et al.} \cite{wang2020few1}&  2020   & $\surd$ & & $\surd$ & &&&$\surd$&Shape morphing\\
			Zhao \emph{et al.} \cite{zhao2021few}&  2021   & $\surd$ & & $\surd$ & &&&$\surd$&Multi-prototype representation\\
			Hao \emph{et al.} \cite{hao20213d}&  2021  & $\surd$ & &  &$\surd$ &&&$\surd$&Meta learner\\
			Huang \emph{et al.} \cite{huang20213d}&  2021  & $\surd$ & &  & &$\surd$&&$\surd$&Meta learner\\\cline{1-10}
			Michele \emph{et al.} \cite{michele2021generative}& 2019  &  &$\surd$ && &  &$\surd$&$\surd$& Zero-shot framework, dataset\\ 
			Liu \emph{et al.} \cite{liu2021segmenting}&  2022  &  &$\surd$ && &  &$\surd$&$\surd$&Dataset, embedding network\\\hline 
		\end{tabular}
		\begin{tablenotes}
			\item[1] Each method is represented by its first word.
			\item[2] “Point” indicates that the sample type is point clouds.
		\end{tablenotes}
	\end{threeparttable}
\end{table*}

\subsection{Few-shot 3D Segmentation}
The majority of available few-shot 3DS  methods counting on the metric learning paradigm have been relatively deeply researched, while a small proportion of them resorting to other tools, such as fine-tune-based or parameter prediction-based strategies, have  been studied rarely. This section focuses on the existing few-shot 3DS  methods and discusses them as follows.

\begin{figure}[!t]
	\centering
	\includegraphics[width=0.9\linewidth]{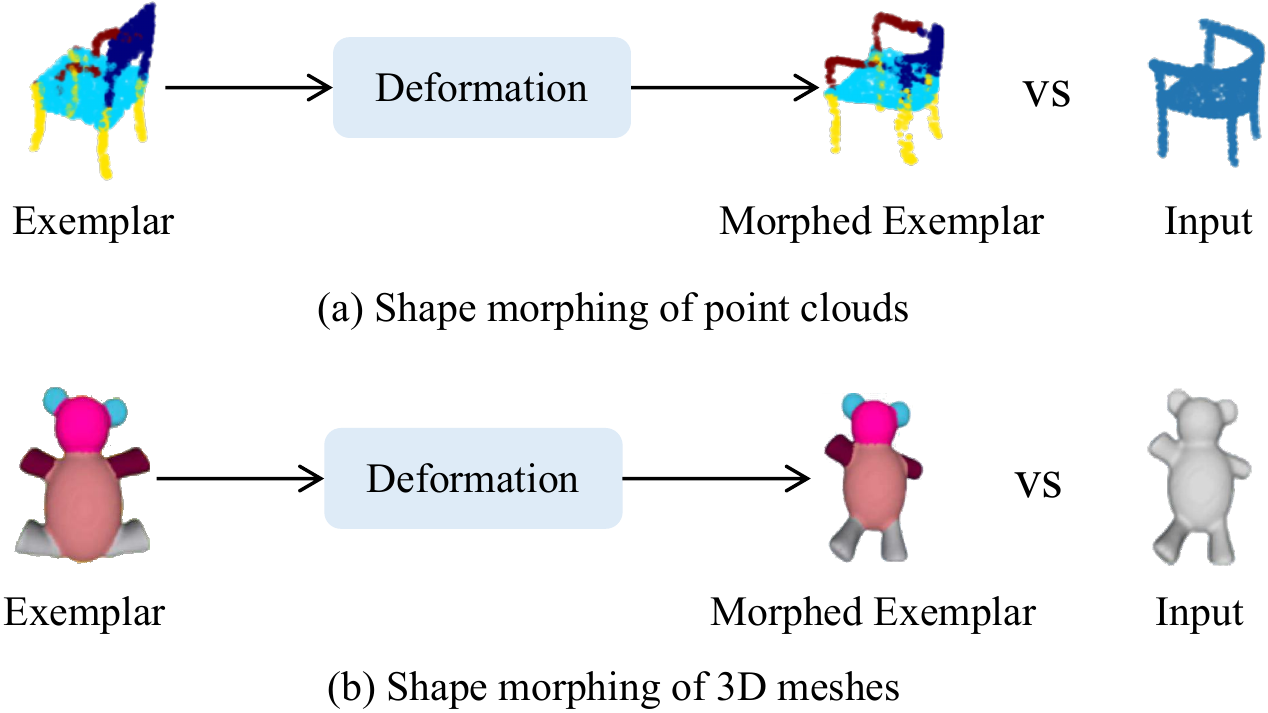}
	\caption{The operation of shape morphing in few-shot 3DS based on metric learning (adapted from \cite{wang2020few1}). The samples are selected from \cite{yuan2020ross}, \cite{wang2020few1}. }
	\label{shape}
\end{figure}

\subsubsection{Metric Learning-based methods}

The methods based on metric learning generally  transfer the labels from the given 3D exemplars to the input ones by calculating the similarity or proximity in a shared space. Dissimilar to few-shot 2D segmentation allocating labels in an embedding space regularly, the approaches related to 3DS also have a propensity to conduct labels assignment in the input space.

From the perspective of taking the embedding space as the public metric space,  prototype networks are also integrated into 3DS for grappling with limited training samples. 
The complicated point clouds in the support set were mapped as multiple prototypes of diverse shapes, which were then applied to measure the similarity with query embeddings in \cite{chen2020compositional}. 
Multiple prototypes were estimated to propagate labels  based on correlations between prototypes and query points \cite{zhao2021few}.

Regarding 3D shape space as the common metric space,  a conventional completion paradigm is to alter the given 3D exemplar into the same shape as the input and then to transfer  labels from the deformed exemplar to the input  for segmentation, as indicated in Fig. \ref{shape}. 
In  \cite{yuan2020ross}, labeled meshes were first morphed to reap shapes that  were the same as inputs.  Then,   a label migration was conducted according to the distance between points sampled from the deformed  and input 3D meshes.  The 3D template was  transformed towards the input shape, which  facilitated the subsequent label alignment operation \cite{wang2020few1}. Moreover, instead of calculating the spatial distance between two points directly,  a probability distribution function was learned to assist the predictions on part-specific labels.

\subsubsection{Parameter Prediction-based Methods} 
Parameter prediction-based methods were also applied to tackle few-shot 3DS.  A potential space associated with diverse 3DS  functions was learned from a significant number of tasks, where the limited training data about unseen shapes was leveraged to dynamically generate a task-specific function for rapid generalization on target shapes in \cite{hao20213d}.

\subsubsection{Fine-tune-based Methods}
There are also some approaches that have recourse to fine-tune policies to cope with few-shot 3DS. In addition to meta-learning \cite{huang20213d}, pre-training paradigms constructed on self-supervised \cite{sharma2019learning,sharma2020self} or unsupervised learning \cite{chen2019bae} are also integrated into few-shot 3DS. 

Under the support of contrastive learning and meta-learning,  pre-training offline was conducted to represent  more discriminative features on shapes and estimate the optimal model initialization in \cite{huang20213d}. 
Based on the initialized parameters, the model could effectively adjust itself  and converge quickly on new tasks with a handful of training samples.  A self-supervised loss function was developed in \cite{sharma2019learning} to assist in the estimation on dimension-fixed feature vectors of 3D points. 
When a small amount of training data was provided, the segmentation model initialized by the pre-trained embedding network could be further fine-tuned for  performance gains. A self-supervised strategy was adopted  in  \cite{sharma2020self} to heuristically represent a point cloud exemplar in a metric space ahead of schedule. The learned point embedding network was then employed to learn the segmentation network  under a few samples.  In \cite{chen2019bae}, an novel autoencoder was designed, where the encoder aimed to represent 3D shape in a feature space, and the branched decoder endeavored to learn a compact representation for each frequently occurring shape. Under this setting, the  model pre-trained in an unsupervised way could quickly acclimate to the segmentation of new 3D shapes by an uncomplicated adjustment with the guidance of only one or a few labeled training data.

\subsection{Zero-shot 3D Segmentation}
Existing zero-shot 3DS  approaches generally adopt generative models to cope with  unseen 3D objects with zero labels, which also follows the learning paradigm illustrated in Fig. \ref{generative}. Similar to generative model-based ISS methods \cite{gu2020context,li2020consistent},  a generator was trained on the seen training data to generate pseudo but semantically consistent features for the unseen point cloud samples  \cite{michele2021generative}. The generated pseudo representations were employed to fine-tune a classifier for overcoming the challenges arising from the absence of unseen point cloud categories and disequilibrium size between seen and unseen samples. Analogous ideas were also adopted to tackle the zero-shot 3D scene segmentation in \cite{liu2021segmenting}, where a regularizer was designed to assist the generation on semantically consistent features of both seen and unseen classes. These synthesized features were then leveraged to adjust the classifier for accurate prediction on unseen objects.

\subsection{Summary}
Table \ref{3d}  sums up the few/zero-shot 3DS  methods reviewed in this paper. It can be discovered that compared with ISS and VOS, the cooperation between few-shot learning (or zero-shot learning) and 3DS is still in its infancy. Due to the disorder and complexity of 3D samples, these technical solutions that show great effectiveness in 2D space are tricky to be  applied in 3D space directly, which limits the development of few/zero-shot 3DS.  How to deal with this challenge effectively with a particular solution is still a promising topic in the future.

\section{discussion}\label{section8}
In spite of the fact that few/zero-shot visual semantic segmentation has achieved a considerable breakthrough and effectively settled the issues caused by a few or even zero annotated samples  in both 2D and 3D space, there are still a variety of obstacles to their applications in real scenes. This section aims to  discuss some open challenges  and list some applications of few/zero-shot visual semantic segmentation.

\textbf{Cross-domain transferability:} Transferability plays a crucial role for the computer vision community \cite{zhang2020autonomous}. For visual semantic segmentation, the transferability is principally reflected in two aspects: crossing different domains and crossing diverse categories, where the cross-category transferability has received extensive attention  in few/zero-shot  visual semantic segmentation. However, the existing  methods are required to follow a solid premise: the samples of unseen classes must have the same distribution as base classes, which places restrictions on the serviceability in real scenarios \cite{boudiaf2021few}. Although visual semantic segmentation methods based on domain adaptation can bridge the inter-domain gap, they fail to generalize to unseen categories effectively. Therefore, integrating domain adaptation with cross-category adaptation seamlessly  (i.e., adopting adversarial training with few-shot learning \cite{zhao2021domain}) 
can be a promising way to significantly boost the performance of few/zero-shot segmentation models in different domains.

\textbf{Generalized few/zero-shot segmentation:} Generalized few/zero-shot visual segmentation requires the model to realize the segmentation on both base and unseen classes. Most of few/zero-shot visual segmentation algorithms tend to deal with the segmentation of unseen categories while ignoring the performance on base classes  in inference. 
For example, the methods based on parameter prediction adjust the classifier weights only for a specified category, which may lead to performance degradation on base classes. 
Furthermore, practical applications also demand that the segmentation model should be able to segment multiple categories of objects, including seen and unseen categories. Therefore, pursuing generalized few/zero-shot segmentation algorithms is one of the valuable future topics.

\textbf{Weakly-supervised few/zero-shot segmentation:} Even though the existing few/zero-shot visual segmentation approaches have greatly alleviated the requirements for annotations of target objects, a large number of base-class samples with  labels are still indispensable \cite{xie2021efficient,seong2021hierarchical,liu2022axial} to learn to separate unseen categories of interest.  
However, it is difficult and challenging to collect such a great deal  of labeled base-class samples in practice. 
The strictness of practical applications entails that few/zero-shot visual semantic segmentation methods are competent to precisely separate unseen categories in a more efficient way. Consequently, combining weakly-supervised learning, or even unsupervised learning \cite{amac2022masksplit}, with few/zero-shot visual semantic segmentation will be a constructive subject in further development.

\textbf{Multi-modal supervision:} Most of few-shot visual semantic segmentation approaches are inclined to rely on insufficient supervision signals from a single-modal support set, which makes them tricky and laborious to deal with the gap between support and query samples effectively.  For example, in few-shot ISS, the type of support samples is all 2D images. However, other modalities, such as language expressions, also  can be used as auxiliary supervision signals in addition to images.  
The zero-shot visual semantic segmentation utilizes the supervisory information from other modalities, such as word embeddings and optical flows, which effectively fills the vacancy in supervision information and conducts reliable predictions for  target categories. 
Therefore, it is  promising for few-shot visual semantic segmentation to  turn to other modalities to augment the supervision information. 
In addition, the utilization of more modalities is also of great significance to further promote the richness of supervision information for zero-shot visual semantic segmentation. 

\textbf{Lightweight network:} Nowadays, embedded devices play a significant role in both civil and military fields, which puts forward higher requirements on the computational overhead and runtime costs. Whereas, the majority of few/zero-shot visual semantic segmentation approaches resort to different backbones, such as VGG16 \cite{simonyan2014very}, or more complicated structure for cross-class prediction, which  requires higher computational costs and memory usage and is disadvantageous for deployments. Moreover, althrough real-time performance can be achieved by decreasing the model size, undesirable performance degradation may be inevitably incurred \cite{zhao2020monocular}. Therefore, how to design a lightweight network with both real-time performance and high-quality prediction is a direction worthy of efforts for few/zero-shot visual semantic segmentation.

\textbf{Cross-task collaboration:} It is natural that the role of a few/zero-shot visual semantic segmentation model is  insufficient to contend with diverse challenges in real scenarios. Therefore, combining visual semantic segmentation with other computer vision tasks  will be a promising direction to expand the abilities of the visual semantic segmentation model in some aspects. 
For instance, 
it is advantageous to leverage few-shot object detection to provide more efficient annotations for weakly-supervised few-shot visual semantic segmentation \cite{bonechi2018generating}. Moreover, collaborating across diverse tasks will also promote cooperation and information sharing between various tasks, promising to maximize the value of limited samples.

\textbf{Cross-task learning:} As summarized above, the technical solutions of ISS, VOS, and 3DS in both few-shot and zero-shot scenarios have some commonalities and universalities, which makes the solutions proposed in one segmentation field can be referenced by another one. For example, on the one hand, some methods \cite{wang2020few,liu2021harmonic}, which are proposed to solve the appearance gap between support  and query samples in few-shot ISS, can be used to deal with  long-time sequences  in few-shot VOS. On the other hand, some strategies \cite{seong2020kernelized,yang2020collaborative}, which  are proposed to address mismatching of similar objects in few-shot VOS, can also be considered in few-shot ISS.

\textbf{Applications:} 
Compared with conventional visual semantic segmentation, few/zero-shot semantic segmentation algorithms have more significant application advantages. Take the petrochemical industry for example. Firstly, few/zero-shot semantic segmentation methods can  be applied in the situations where labeled training data is scarce, such as oil and gas pipeline leakage detection \cite{lu2021effective}. Secondly, few/zero-shot visual-semantic segmentation provides a flexible solution for fast cross-class adaptation, which plays an important  role in addressing diverse types of pipeline defects, such as deformation, corrosion, scaling and cracks \cite{ravishankar2022darts}.  Thirdly, combining the limited number of cross-modal samples (i.e., infrared images) and breaking through the barriers of  multi-modal  few/zero-shot semantic segmentation methods can advance the applications in all-weather  scenarios, such as safety production monitoring of petroleum and petrochemical products \cite{zhou2021gmnet,patil2022multi}. 
In a word, employing  few/zero-shot semantic segmentation in real scenarios, such as petrochemical industry, plays a positive role in promoting the production and life to become more  intelligent and autonomous.

\section{Conclusion}
This paper focuses on the applications of few-shot learning on visual semantic segmentation. To this end, we perform an exhaustive and systematic survey on related works of three typical few/zero-shot segmentation tasks, including few/zero-shot  ISS, few/zero-shot  VOS, and few/zero-shot 3DS. Moreover,  we explore the commonnalities and discrepancies of few/zero-shot settlements under different segmentation circumstances. 
Besides, we analyze these existing few/zero-shot segmentation methods and list some challenges and valuable directions for follow-up researchers.


%

%
%

\ifCLASSOPTIONcaptionsoff
  \newpage
\fi




%
\bibliographystyle{bibtex/IEEEtran.bst} 
\bibliography{ref} 

%

\begin{IEEEbiography}[{\includegraphics[width=1in,height=1.25in,clip,keepaspectratio]{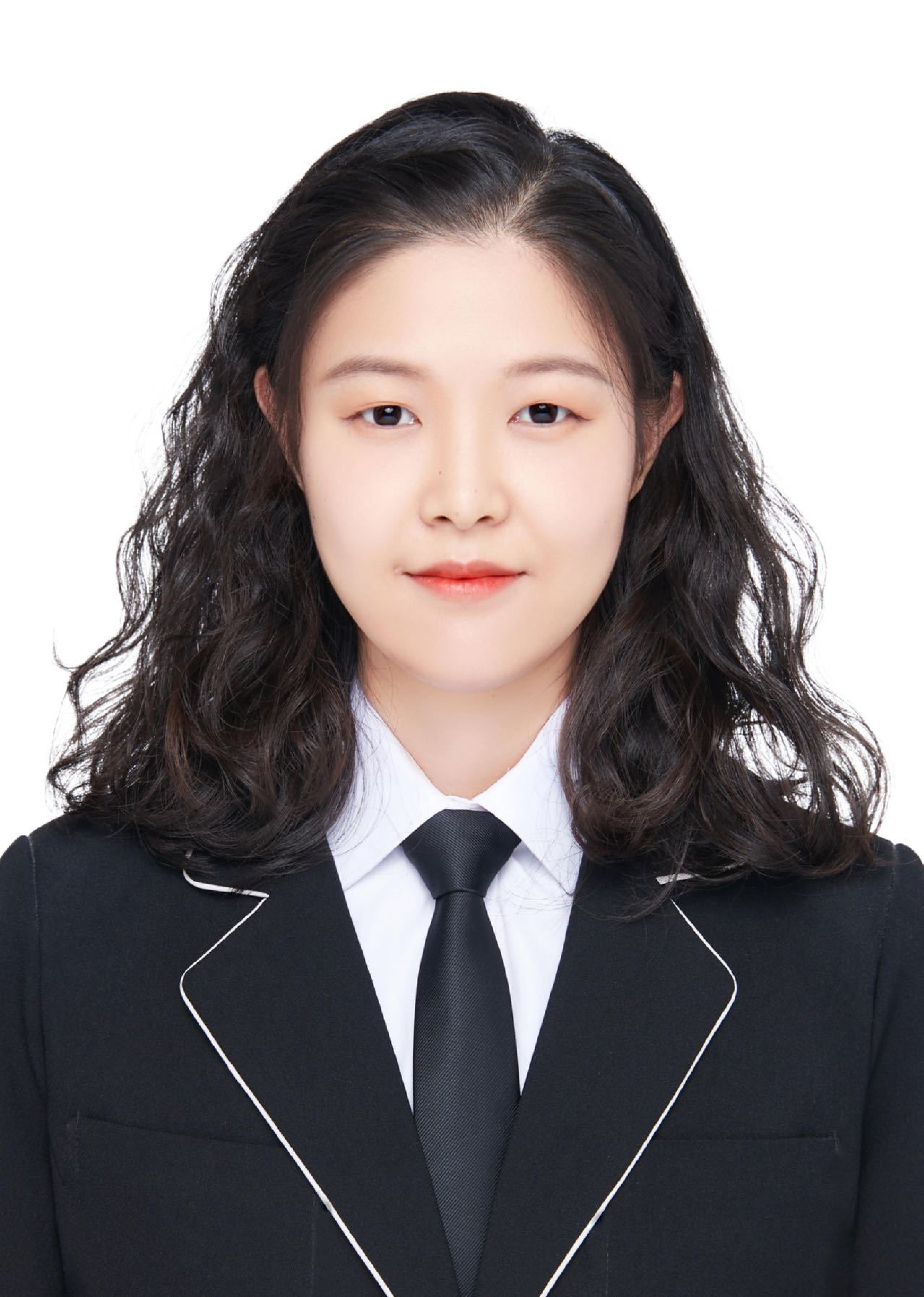}}]{Wenqi Ren}
received the B.S. degree in electrical engineering and automation, from the East China University of Science and Technology, Shanghai, China, in 2020, where she is currently working toward the Ph.D. degree in control science and engineering. Her research interests include meta-learning, few-shot learning, domain adaptation, and scene understanding.
\end{IEEEbiography}

\begin{IEEEbiography}[{\includegraphics[width=1in,height=1.25in,clip,keepaspectratio]{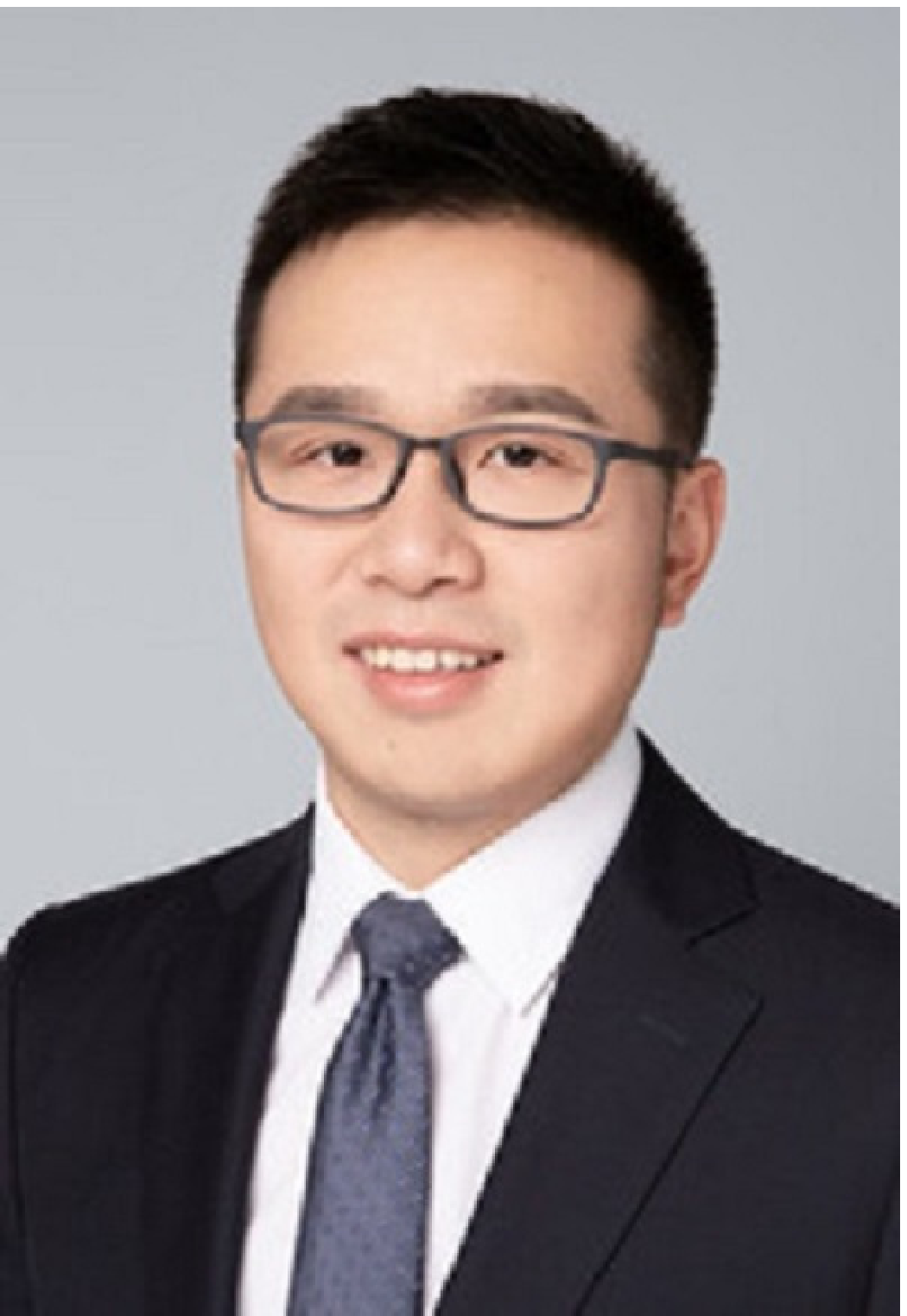}}]{Yang Tang}
(Senior Member, IEEE) received the B.S. and Ph.D. degrees in electrical engineering from Donghua University in 2006 and 2010, respectively. From 2008 to 2010, he was a Research Associate with The Hong Kong Polytechnic University, China. From 2011 to 2015, he was a Post-Doctoral Researcher with the Humboldt University of Berlin, Germany, and with the Potsdam Institute for Climate Impact Research, Germany. He is now a Professor with the East China University of Science and Technology. His current research interests include distributed estimation/control/optimization, cyber-physical systems, hybrid dynamical systems, computer vision, reinforcement learning and their applications. 

He was a recipient of the Alexander von Humboldt Fellowship and has been the ISI Highly Cited Researchers Award by Clarivate Analytics from 2017. He is a Senior Board Member of Scientific Reports, an Associate Editor of IEEE Transactions on Neural Networks and Learning Systems, IEEE Transactions on Cybernetics, IEEE Transactions on Circuits and Systems-I: Regular Papers, IEEE Transactions on Emerging Topics in Computational Intelligence, IEEE Systems Journal and Engineering Applications of Artificial Intelligence (IFAC Journal), etc. He is a Leading Guest Editor for special issues in IEEE Transactions on Emerging Topics in Computational Intelligence and IEEE Transactions on Cognitive and Developmental Systems.
\end{IEEEbiography}

\begin{IEEEbiography}[{\includegraphics[width=1in,height=1.25in,clip,keepaspectratio]{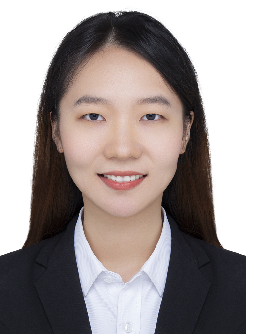}}]{Qiyu Sun}
received the B.S. degree in automation, from the East China University of Science and Technology, Shanghai, China, in 2019, where she is currently working toward the Ph.D. degree in control science and engineering. Her research interests include 3D scene understanding, domain adaptation, and deep learning.
\end{IEEEbiography}

\begin{IEEEbiography}[{\includegraphics[width=1in,height=1.25in,clip,keepaspectratio]{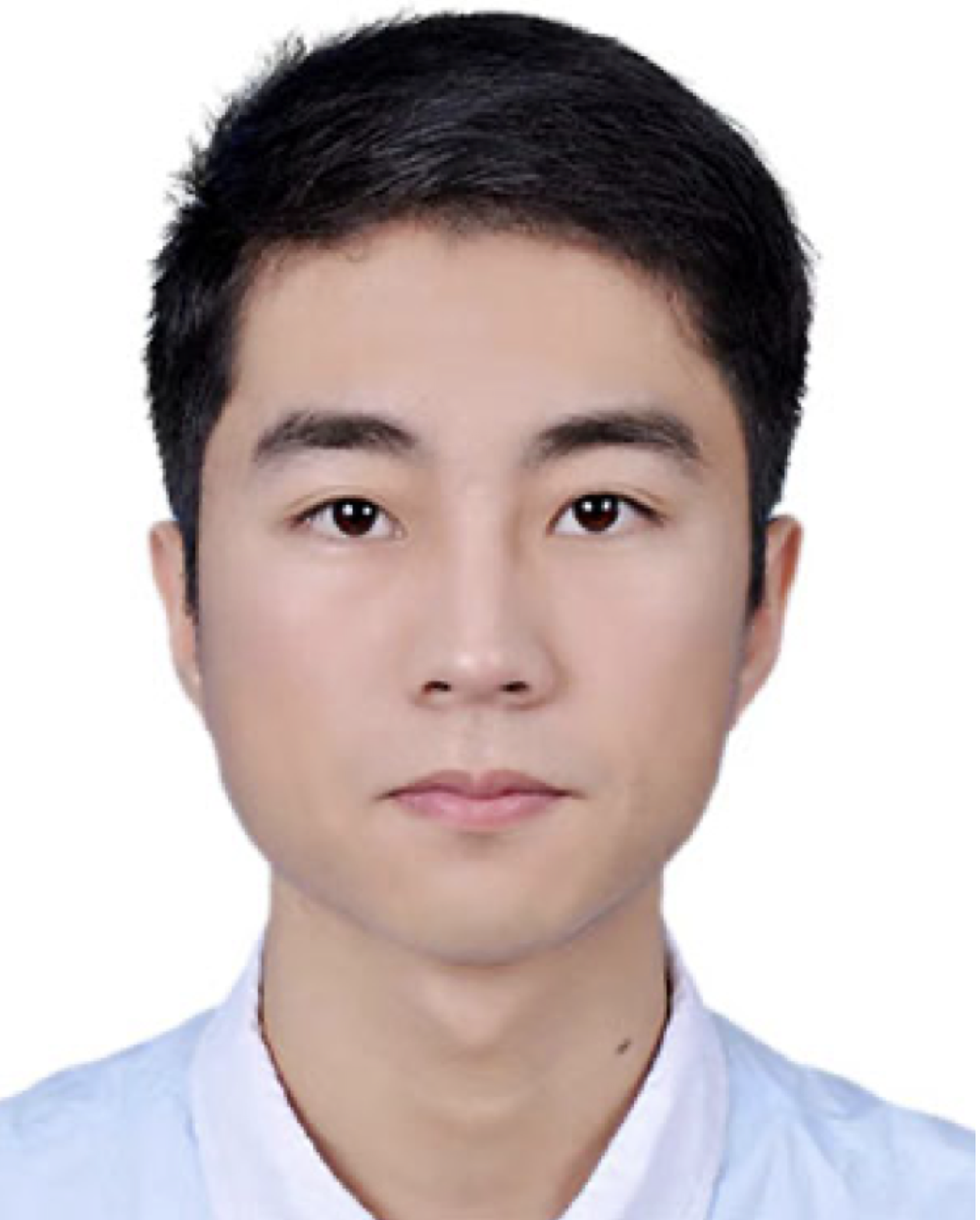}}]{Chaoqiang Zhao}
	received the B.S. degree in automation from the East China University of Science and Technology, Shanghai, China, in 2018, where he is currently working toward the Ph.D. degree in control science and engineering. His research interests include depth perception, visual odometry, and stereo vision.
\end{IEEEbiography}

\begin{IEEEbiography}[{\includegraphics*[width=1in,height=1.25in,clip,keepaspectratio]{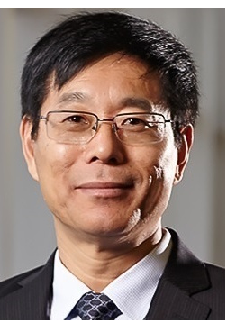}}]{Qing-Long Han} (Fellow, IEEE) received the B.Sc. degree in Mathematics 
	from Shandong Normal University, Jinan, China, in 1983, and the M.Sc. and Ph.D. degrees in Control Engineering from East China University of Science and Technology, 
	Shanghai, China, in 1992 and 1997, respectively.
	
	Professor Han is Pro Vice-Chancellor (Research Quality) and a Distinguished Professor at Swinburne University of Technology, Melbourne, Australia. He held various academic 
	and management positions at Griffith University and Central Queensland University, Australia. His research interests include networked control systems, multi-agent systems, 
	time-delay systems, smart grids, unmanned surface vehicles, and neural networks.
	
	Professor Han was awarded The 2021 Norbert Wiener Award (the Highest Award in systems science and engineering, and cybernetics) and The 2021 M. A. Sargent Medal 
	(the Highest Award of the Electrical College Board of Engineers Australia). He was the recipient of The 2022 IEEE SMC Society Andrew P. Sage Best Transactions Paper Award, 
	The 2021 IEEE/CAA Journal of Automatica Sinica Norbert Wiener Review Award, The 2020 IEEE Systems, Man, and Cybernetics Society Andrew P. Sage Best Transactions Paper Award, The 2020 IEEE Transactions on Industrial Informatics Outstanding Paper Award, 
	and The 2019 IEEE Systems, Man, and Cybernetics Society Society Andrew P. Sage Best Transactions Paper Award.
	
	Professor Han is a Member of the Academia Europaea (The Academy of Europe). He is a Fellow of The International Federation of Automatic Control (IFAC) and a Fellow of The Institution of Engineers Australia (IEAust).  
	He is a Highly Cited Researcher  in both Engineering and Computer Science (Clarivate Analytics). He has served as an AdCom Member of IEEE Industrial Electronics Society (IES), a Member of IEEE IES Fellows Committee, 
	and Chair of IEEE IES Technical Committee on Networked Control Systems. Currently, he is Co-Editor-in-Chief of IEEE Transactions on Industrial Informatics, Deputy Editor-in-Chief of IEEE/CAA JOURNAL OF AUTOMATICA SINICA, Co-Editor of Australian Journal of Electrical and Electronic Engineering, 
	an Associate Editor for 12 international journals, including the IEEE TRANSACTIONS ON CYBERNETICS, IEEE INDUSTRIAL ELECTRONICS MAGAZINE, Control Engineering Practice, and Information Sciences, and a Guest Editor for 14 Special Issues.
\end{IEEEbiography}

%
%
%




\end{document}